\documentclass[conference]{IEEEtran}
\IEEEoverridecommandlockouts

\usepackage{cite}
\usepackage{amsmath,amsthm,amssymb,amsfonts}
\usepackage{graphicx}
\usepackage{textcomp}
\usepackage{xcolor}
\def\BibTeX{{\rm B\kern-.05em{\sc i\kern-.025em b}\kern-.08em
    T\kern-.1667em\lower.7ex\hbox{E}\kern-.125emX}}

\usepackage{bm}
\usepackage{graphicx}
\usepackage{subcaption}
\usepackage{textcomp}
\usepackage{booktabs}
\usepackage{nicefrac}
\usepackage{multirow}
\usepackage{amsfonts}
\usepackage{lipsum}
\usepackage{float}
\usepackage{algorithm} 
\usepackage{algpseudocode} 
\usepackage[acronym]{glossaries}
\usepackage{tikzit}
\usepackage{wrapfig}
\usepackage{nicefrac}
\usepackage{fontawesome}
\graphicspath{{./Figures}}
\usepackage[hidelinks]{hyperref}
\usepackage{comment}

\usepackage[acronym]{glossaries}

\newacronym{gd}{GD}{Gradient Descent}
\newacronym{ot}{OT}{Optimal Transport}
\newacronym{dr}{DR}{Dimensionality Reduction}
\newacronym{gw}{GW}{Gromov Wasserstein}
\newacronym{mds}{MDS}{Multidimensional Scaling}
\newacronym{isomap}{ISOMAP}{Isometric Mapping}
\newacronym{gw-mds}{GW-MDS}{Gromov Wasserstein MDS}
\newacronym{pca}{PCA}{Principal Component Analysis}
\newacronym{ewca}{EWCA}{Entropic Wasserstein Component Analysis}

\title{A dimensionality reduction technique based on the Gromov-Wasserstein distance}

\author{\IEEEauthorblockN{Rafael Pereira Eufrazio$^{1,2}$ \quad Eduardo Fernandes Montesuma$^{3}$ \quad Charles Casimiro Cavalcante$^{2}$}
\IEEEauthorblockA{$^{1}$Instituto Federal de Educação, Ciência e Tecnologia do Ceará, Canindé-CE, Brazil\\
$^{2}$Federal University of Ceara, Fortaleza-CE, Brazil\\
$^{3}$Université Paris-Saclay, CEA, List, F-91120 Palaiseau, France}}

\newcommand{\argmin}[1]{\underset{#1}{\text{argmin}}\,}

\definecolor{myorange}{rgb}{0.906,0.435,0.317}
\definecolor{myblue}{rgb}{0.0,0.314,0.408}
\definecolor{private}{rgb}{0.5,0.0,0.0}
\definecolor{public}{rgb}{0.17,0.62,0.17}

\theoremstyle{definition}

\begin{document}
%
\maketitle

\begin{abstract}
Analyzing relationships between objects is a pivotal problem within data science. In this context, Dimensionality reduction (DR) techniques are employed to generate smaller and more manageable data representations. This paper proposes a new method for dimensionality reduction, based on optimal transportation theory and the Gromov-Wasserstein distance. We offer a new probabilistic view of the classical Multidimensional Scaling (MDS) algorithm and the nonlinear dimensionality reduction algorithm, Isomap (Isometric Mapping or Isometric Feature Mapping) that extends the classical MDS, in which we use the Gromov-Wasserstein distance between the probability measure of high-dimensional data, and its low-dimensional representation. Through gradient descent, our method embeds high-dimensional data into a lower-dimensional space, providing a robust and efficient solution for analyzing complex high-dimensional datasets.
\end{abstract}
\section{Introduction}\label{sec:intro}

Analyzing relationships between objects is a pivotal problem within data science. In this context, \gls{mds} is a technique for representing these objects, in a low dimensional space, based on the degree of similarity, or dissimilarity, between these objects in their original space~\cite{borg2007modern}. As such, this method belongs to the wider class of techniques known as \gls{dr}, a problem within unsupervised learning and machine learning. In this context, low dimensional representations of data offer numerous advantages, such as improved pattern recognition and structure identification, as well as faster processing for downstream tasks. We refer readers to~\cite{lee2007nonlinear}, for a review of \gls{dr} algorithms and principles.

\gls{dr} algorithms create a low dimensional representation $\mathbf{Y}$ for high dimensional data $\mathbf{X}$. These methods are divided into two categories, namely, linear, and non-linear methods~\cite{lee2007nonlinear}. Linear methods work via projection, i.e., one devises a matrix $\mathbf{W} \in \mathbb{R}^{p \times d}$, such that $\mathbf{Y} = \mathbf{XW}$. A famous example is \gls{pca}, which projects $\mathbf{X}$ in the direction of its eigenvectors. Non-linear methods work under different principles. For instance, \gls{mds}~\cite{borg2007modern} define $\mathbf{Y}$ such that the pairwise Euclidean distances in high dimensions are preserved. In the context of \gls{mds}, representations are defined in terms of the \emph{stress}, a metric of how much these representations respect the dissimilarity between the original objects. However, this metric does not consider the potential relationship between points at a local level. To remedy this issue, we consider the \gls{gw} distance~\cite{memoli2011gromov}, a metric defined in terms of optimal transportation theory~\cite{villani2009optimal}. In this sense, we provide a probabilistic view of \gls{mds} and the algorithm that extends classical MDS, Isomap (Isometric Mapping or Isometric Feature Mapping) is another nonlinear dimensionality reduction method that preserves geodesic distances, rather than direct Euclidean distances. This approach is particularly effective for data that are on low-dimensional manifolds, capturing local relationships more faithfully.

Recently, in~\cite{van2022probabilistic},~\cite{van2024snekhorn}, and~\cite{van2008visualizing}, different authors have analyzed the \gls{dr} problem through probabilistic lens. In this sense, one assumes some form for the underlying probability measure of high dimensional data. The low dimensional representation is thus optimized to match such measure. In this context, \gls{ot} is a natural tool for comparing, and manipulating probability measures~\cite{montesuma2023recent}. A major challenge in \gls{dr}, is the fact that the probability measure associated with $\mathbf{X}$ is supported in a different space than that of $\mathbf{Y}$. As a result, a natural candidate for comparing these objects is the \gls{gw} distance~\cite{memoli2011gromov}. In this paper, we introduce a practical algorithm for performing \gls{mds} based on the \gls{gw} metric.

We summarize our contributions as follows. First, we provide a new probabilistic view of the classical MDS problem and the Isomap algorithm. This novel formulation has the advantage of capturing local relationships between objects through an optimal transport plan. Second, we provide a new and practical algorithm based on \gls{gd} on the \gls{gw} metric for finding the embeddings of high dimensional objects. As we demonstrate through our experiments, this new formulation produces embeddings whose distances better correlate with those between high dimensional data (Table~\ref{tab:quant_analysisI}) and (Table~\ref{tab:quant_analysisII}).




The rest of this paper is organized as follows. Section~\ref{sec:background} provides an introduction to optimal transport and dimensionality reduction. Section~\ref{sec:methodology} discusses our proposed method. Section~\ref{sec:experiments} shows our experiments in dimensionality reduction. Finally, section~\ref{sec:conclusion} concludes this paper.


\section{Background}\label{sec:background}

\subsection{Dimensionality Reduction}

\gls{dr} is an essential technique in unsupervised machine learning, used to represent high-dimensional data in a more interpretable and manageable form. Given a dataset \mbox{\( X = (x_1, ..., x_n)^\top \in \mathbb{R}^{n \times p} \)}, these techniques seek to construct a low-dimensional representation \( Y = (y_1, ..., y_n)^\top \in \mathbb{R}^{n \times d} \), where \( d < p \). In this paper, we are particularly interested in methods that optimize \( Y \) so that a similarity matrix in the output space corresponds to the similarity matrix in the input space, \( C_X \), according to a specific loss criterion.

We focus on a new formulation of the metric \gls{mds} algorithm~\cite{borg2007modern}. Given a matrix $C_{X} \in \mathbb{R}^{n\times n}$, the goal of this algorithm is in defining $Y$ such that $C_{Y,i,j} = d_{\mathcal{Y}}(y_{i},y_{j})$ preserves $C_{X,i,j} = d_{\mathcal{X}}(x_{i},x_{j})$. In mathematical terms,
\begin{align}
    Y^{\star} = \argmin{y_{1},\cdots,y_{n}} \sum_{i < j}(d_{\mathcal{X}}(x_{i},x_{j}) - d_{\mathcal{Y}}(y_{i},y_{j}))^{2},\label{eq:mds}
\end{align}
where the summation in called \emph{stress}, $\sigma(y_{1},\cdots,y_{n})$. 

An important extension of metric MDS that focuses on preserving geodesic distances, rather than pairwise distances in the input space, is the Isomap algorithm.

Isomap is a nonlinear dimensionality reduction algorithm that extends classical MDS to attempt to preserve the “geodesic” distances between points in a dataset. In general terms, it works as follows~\cite{tenenbaum2000global}:

\begin{itemize}
    \item Construction of the neighborhood graph;
    \item Calculation of geodesic distances;
    \item Dimensionality reduction via classical MDS.
\end{itemize}

By focusing on preserving geodesic rather than Euclidean distances, Isomap is particularly useful in scenarios where the data exhibit strong nonlinear relationships. Unlike purely linear methods such as PCA or traditional MDS, which rely on straight-line distances in the ambient space, Isomap captures local and global manifold structures more accurately. This makes it especially valuable for tasks where the data are believed to reside on smooth, potentially high-curvature surfaces.

\subsection{Optimal Transport}

In this section, we provide a brief overview of \gls{ot}. We refer readers to~\cite{peyre2019computational} for a broader view on the subject, and~\cite{montesuma2023recent} for a review of its applications to machine learning. Let $\mu$ and $\nu$ be two probability measures, and $\{x_{i}\}_{i=1}^{n}$, $\{y_{j}\}_{j=1}^{m}$ be two i.i.d. samples of size $n$ and $m$, respectively. The discrete Kantorovich formulation of \gls{ot} is a linear program,
\begin{align}
    \pi^{\star} = \argmin{\pi \in \Pi(\hat{\mu},\hat{\nu})}\sum_{i=1}^{n}\sum_{j=1}^{m}\pi_{i,j}C_{ij},\label{eq:kantorovich_ot}
\end{align}
where $C_{ij}$ is the ground-cost matrix, which measures the cost of moving $x_{i}$ to $y_{j}$, and $\Pi(\hat{\mu},\hat{\nu}) = \{\pi \in \mathbb{R}^{n\times m}_{+}: \sum_{i}\pi_{ij}=m^{-1}\text{ and } \sum_{j}\pi_{ij}=n^{-1}\}$ is the set of admissible transport plans. Here, $\hat{\mu}$ (resp. $\hat{\nu}$) is the empirical measure $\hat{\mu} = n^{-1}\sum_{i=1}^{n}\delta_{x_{i}}$. When $C_{ij} = d(x_{i},y_{j})$, for a metric $d$, equation~(\ref{eq:kantorovich_ot}) defines a distance between probability measures,
\begin{align}
    W_{p}(\hat{\mu},\hat{\nu})^{p} = \sum_{i=1}^{n}\sum_{j=1}^{m}\pi^{\star}_{i,j}d(x_{i},y_{j})^{p}.\label{eq:wasserstein_distance}
\end{align}
A common choice is $p=2$, and $d(x_{i},y_{j}) = \lVert x_{i}-y_{j} \rVert_{2}$. A major limitation of equations~(\ref{eq:kantorovich_ot}) and~(\ref{eq:wasserstein_distance}), is that it presupposes that $x_{i}$ and $y_{j}$ live in the same ambient space, so that distances can be computed. This motivated~\cite{memoli2011gromov} to propose the \gls{gw} formulation of \gls{ot}, when $\mu$ and $\nu$ live in \emph{incomparable spaces},
\begin{equation}
    \pi^{\star} = \argmin{\pi \in \Pi(\hat{\mu},\hat{\nu})}\sum_{i,j,k,\ell}(d_{\mathcal{X}}(x_{i},x_{j}) - d_{\mathcal{Y}}(y_{k},y_{\ell}))^{2}\pi_{i,j}\pi_{k,\ell},\label{eq:gromov-wasserstein}
\end{equation}
where, one assumes $x_{i} \in \mathcal{X}$ and $y_{j} \in \mathcal{Y}$, and $d_{\mathcal{X}}$ and $d_{\mathcal{Y}}$ are metrics on these spaces. The problem in equation~(\ref{eq:wasserstein_distance}) is quadratic, and like the original \gls{ot} problem, defines a metric between measures $\hat{\mu}$ and $\hat{\nu}$ given by,
\begin{equation}
    GW(\hat{\mu},\hat{\nu}) = \sum_{i,j,k,\ell}(d_{\mathcal{X}}(x_{i},x_{j}) - d_{\mathcal{Y}}(y_{k},y_{\ell}))^{2}\pi_{i,j}^{\star}\pi_{k,\ell}^{\star}\label{eq:gromov_wasserstein}
\end{equation}

One should compare equations~\ref{eq:gromov_wasserstein} and~\ref{eq:mds}. Note that, while equation~\ref{eq:mds} compares the distances between pairs $(i, j)$ with $i < j$, equation~\ref{eq:gromov_wasserstein} compares all distances $(i,j,k,\ell)$. However, since the transport plan matrix is \emph{sparse} (see, e.g., the discussion in \cite[Chapter 4]{peyre2019computational}), this boils down to a handful of non-zero elements of $\pi^{\star}$. Naturally, since $\pi^{\star}$ is determined via linear programming, it captures the local relationship between objects, rather than comparing all possible $(i, j)$ as in equation~\ref{eq:mds}.

The theoretical underpinnings of Optimal Transport and its Gromov-Wasserstein variant provide a powerful framework to compare distributions defined on potentially different spaces. Building upon these ideas, in the next section, we present our GW-MDS method, which merges concepts from MDS and Gromov-Wasserstein to effectively handle data relationships.

\section{Gromov Wasserstein Multidimensional Scaling}\label{sec:methodology}

Our proposed technique, called \gls{gw-mds}, leverages optimal transport for extending metric \gls{mds}. Our main idea comes from the similarity between the stress in equation~(\ref{eq:mds}), and the Gromov-Wasserstein distance in equation~(\ref{eq:gromov_wasserstein}). As a result, we propose a novel optimization problem denoted by,
\begin{align}
    Y^{\star} = \argmin{y_{1},\cdots,y_{n}}GW(\hat{\mu}, \hat{\nu}),\label{eq:gw_mds}
\end{align}
where $\hat{\mu} = n^{-1}\sum_{i=1}^{n}\delta_{x_{i}}$, and $\hat{\nu} = n^{-1}\sum_{j=1}^{n}\delta_{y_{j}}$. From a theoretical perspective, we embed the low dimensional points into a Wasserstein space through $Y \mapsto \hat{\nu}$. In this sense, we give a probabilistic sense to the original \gls{mds} problem.
\begin{figure*}[ht]
    \centering
    \begin{subfigure}{0.24\linewidth}
        \includegraphics[width=\linewidth]{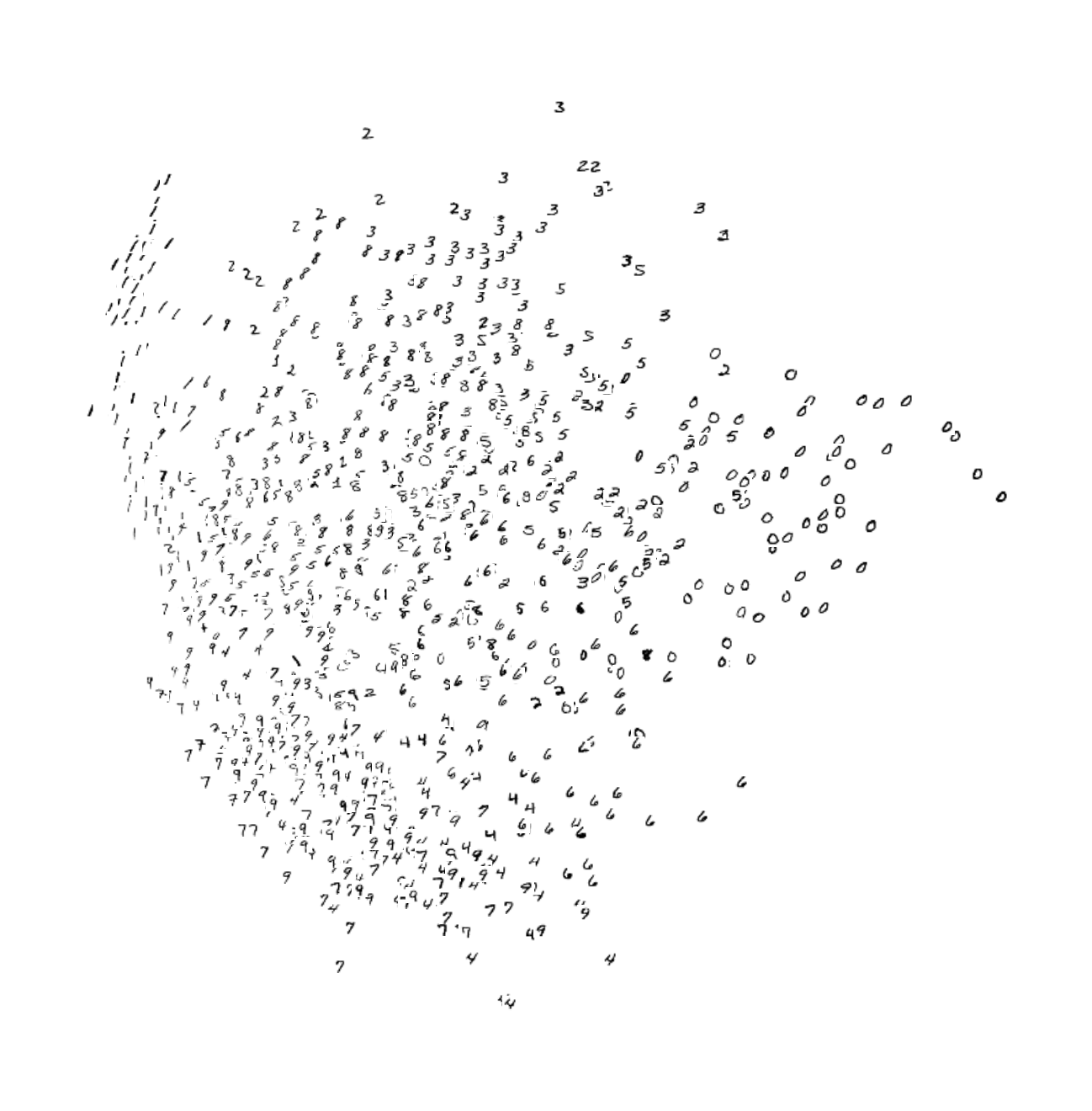}
        \caption{PCA}
    \end{subfigure}\hfill
    \begin{subfigure}{0.24\linewidth}
        \includegraphics[width=\linewidth]{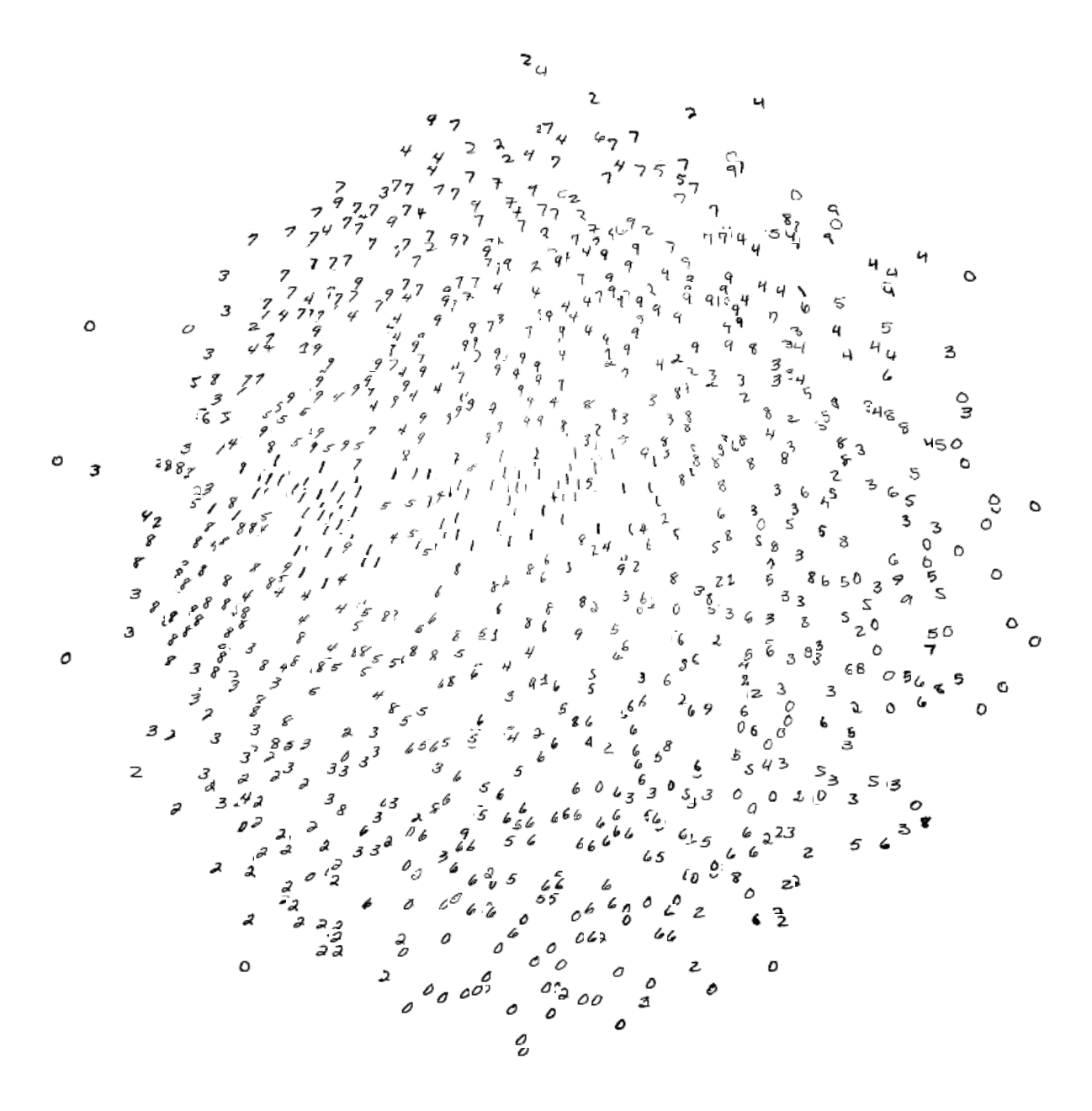}
        \caption{MDS}
    \end{subfigure}\hfill
    \begin{subfigure}{0.24\linewidth}
        \includegraphics[width=\linewidth]{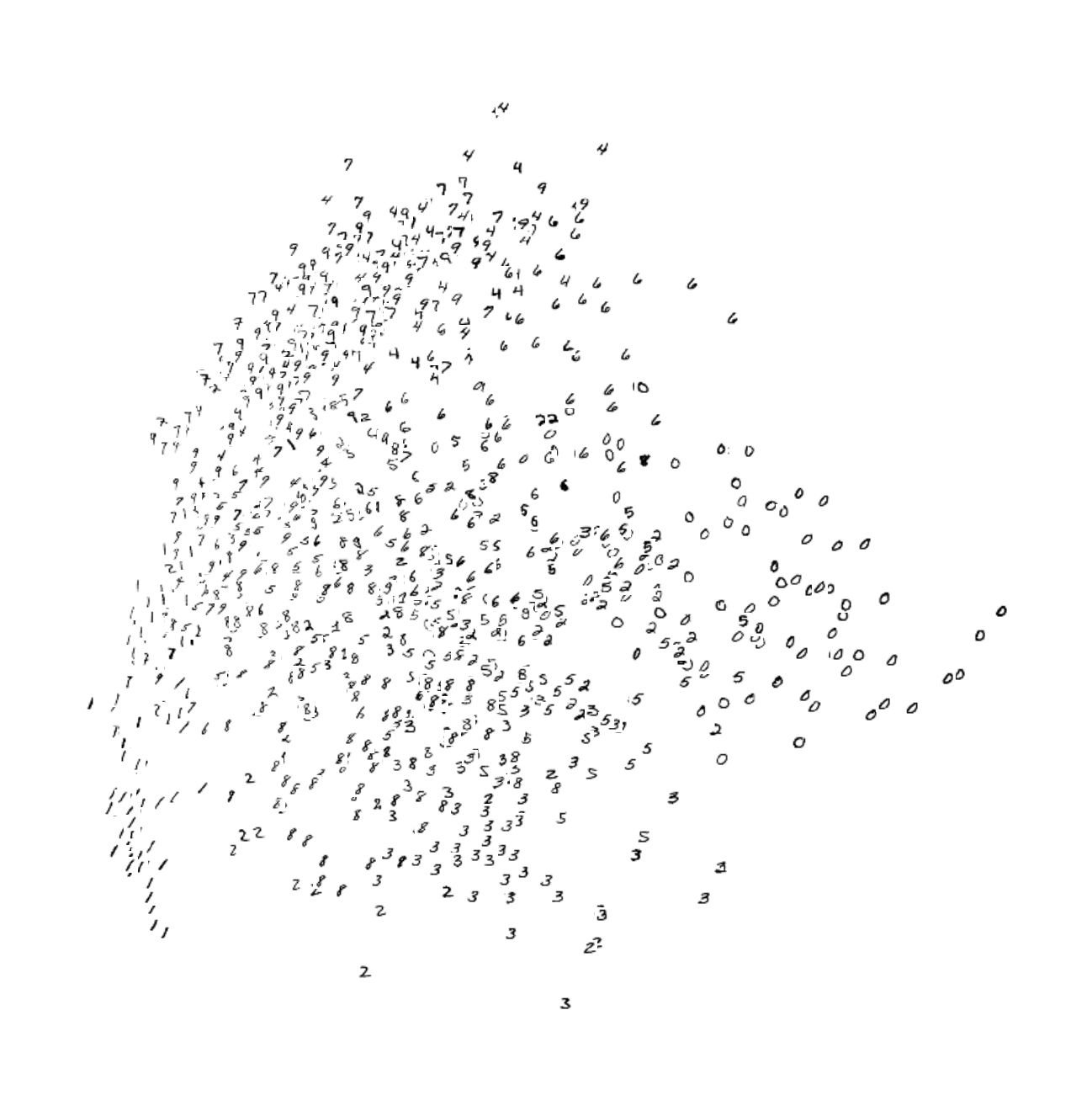}
        \caption{EWCA}
    \end{subfigure}\hfill
    \begin{subfigure}{0.24\linewidth}
        \includegraphics[width=\linewidth]{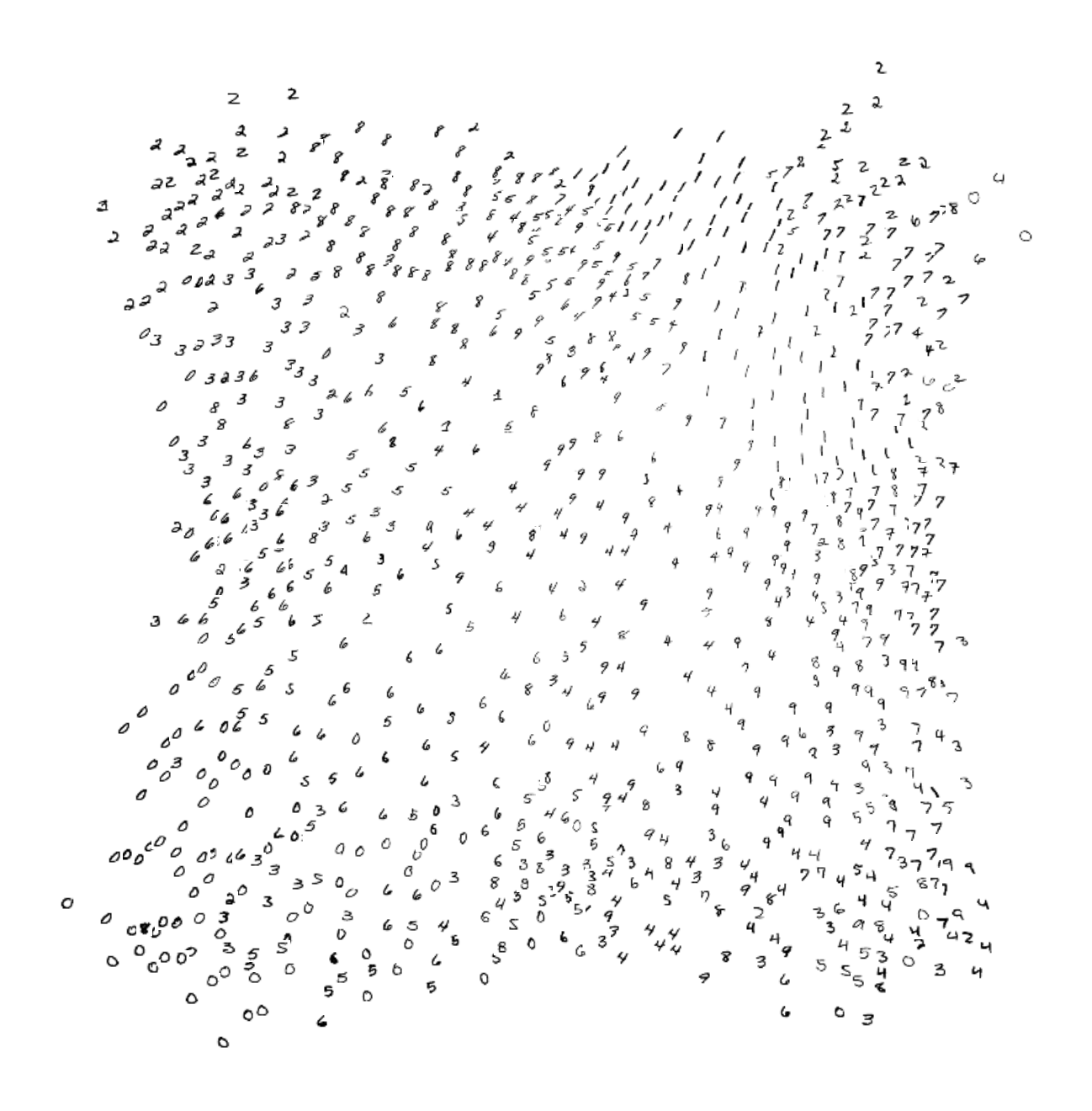}
        \caption{GW-MDS}
    \end{subfigure}
    \begin{subfigure}{0.24\linewidth}
        \includegraphics[width=\linewidth]{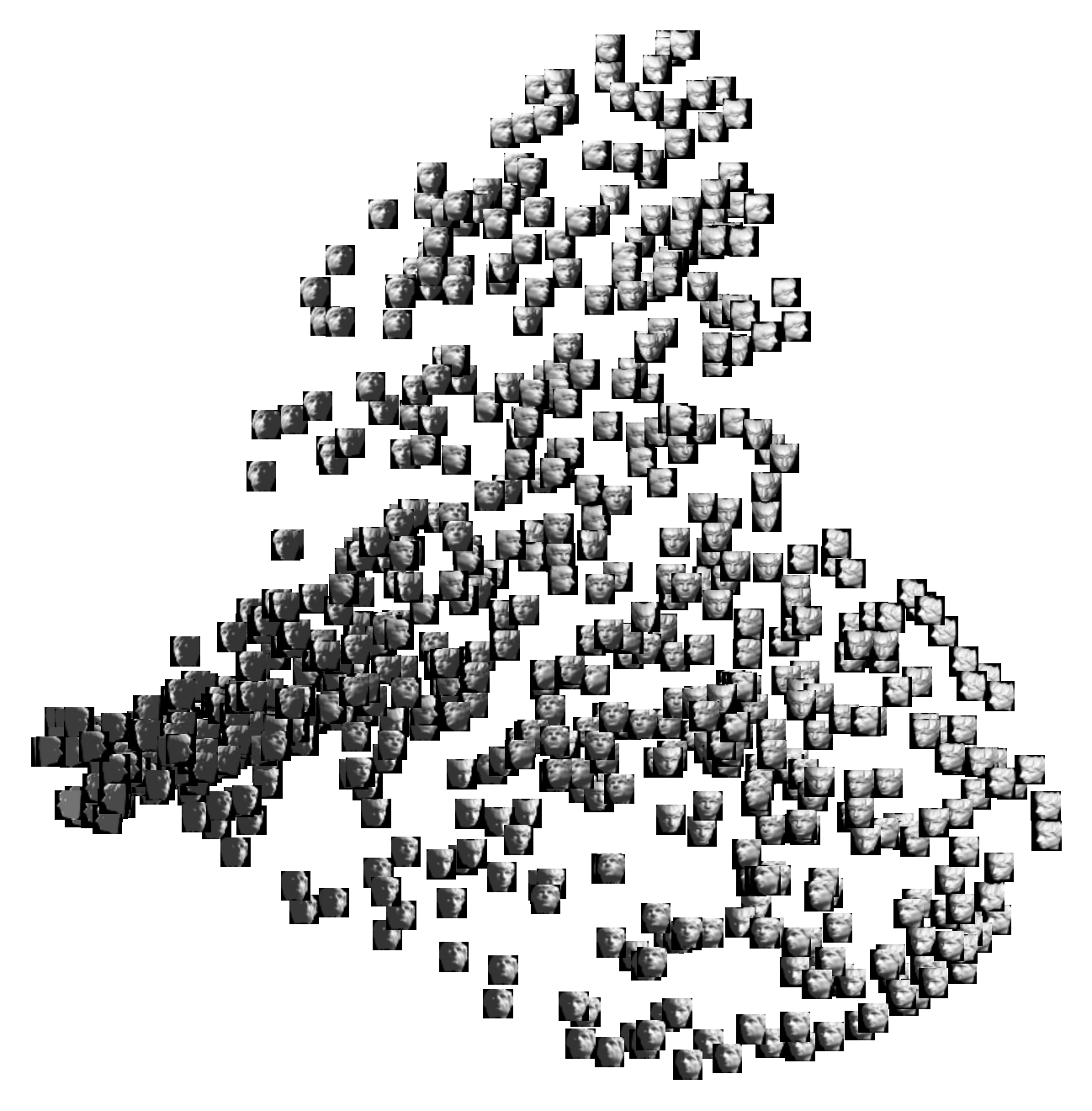}
        \caption{PCA}
    \end{subfigure}\hfill
    \begin{subfigure}{0.24\linewidth}
        \includegraphics[width=\linewidth]{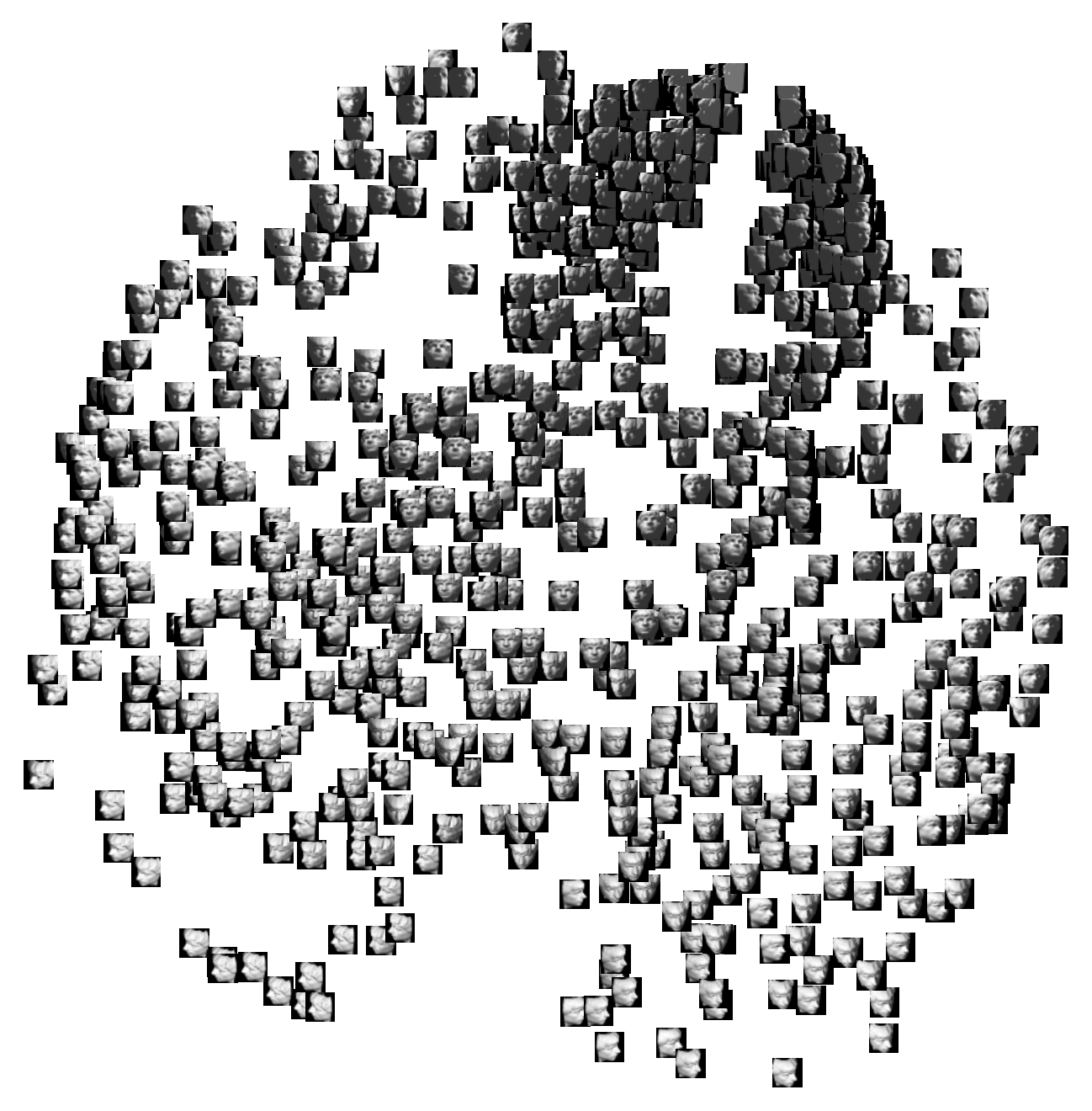}
        \caption{MDS}
    \end{subfigure}\hfill
    \begin{subfigure}{0.24\linewidth}
        \includegraphics[width=\linewidth]{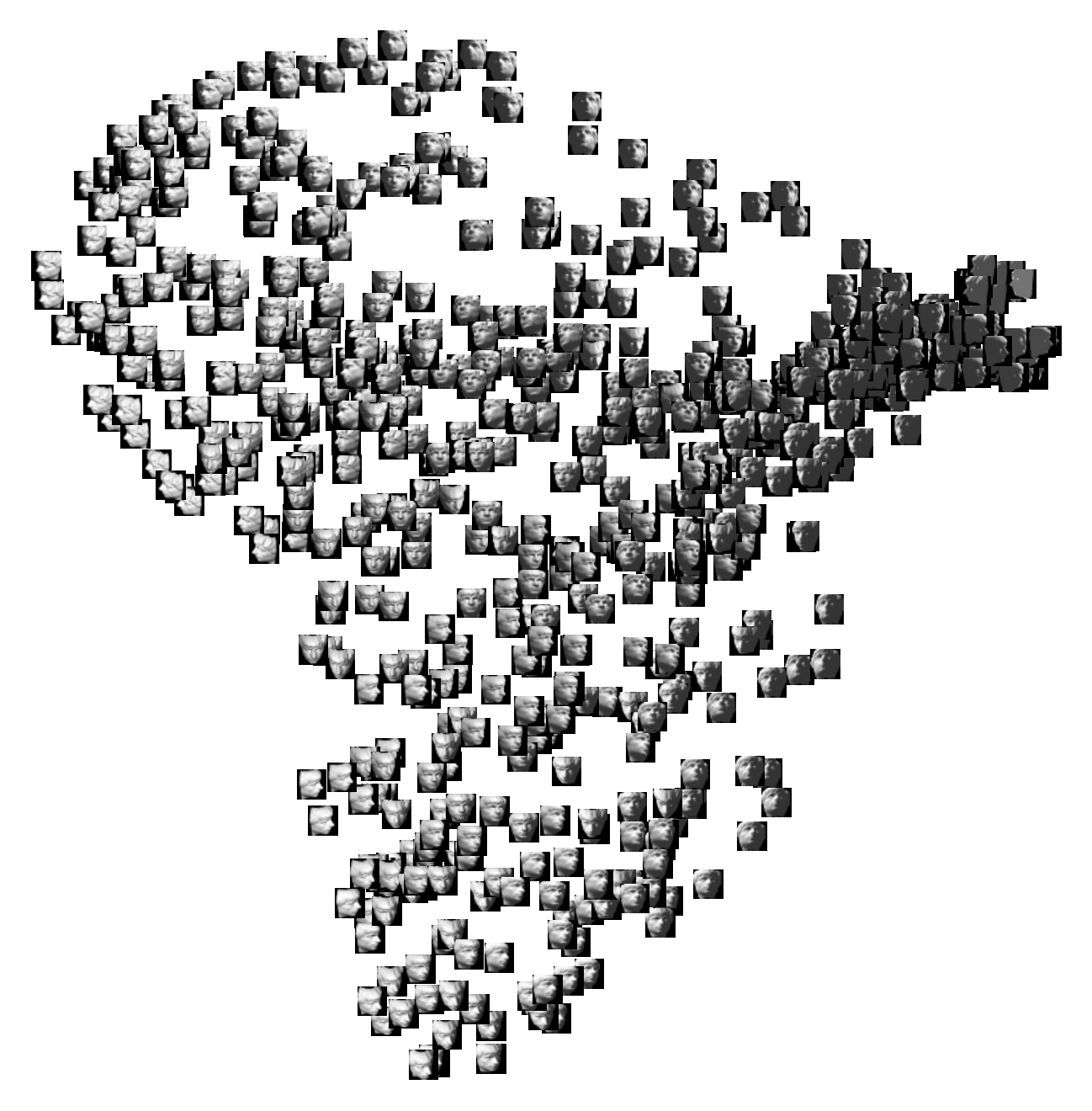}
        \caption{EWCA}
    \end{subfigure}\hfill
    \begin{subfigure}{0.24\linewidth}
        \includegraphics[width=\linewidth]{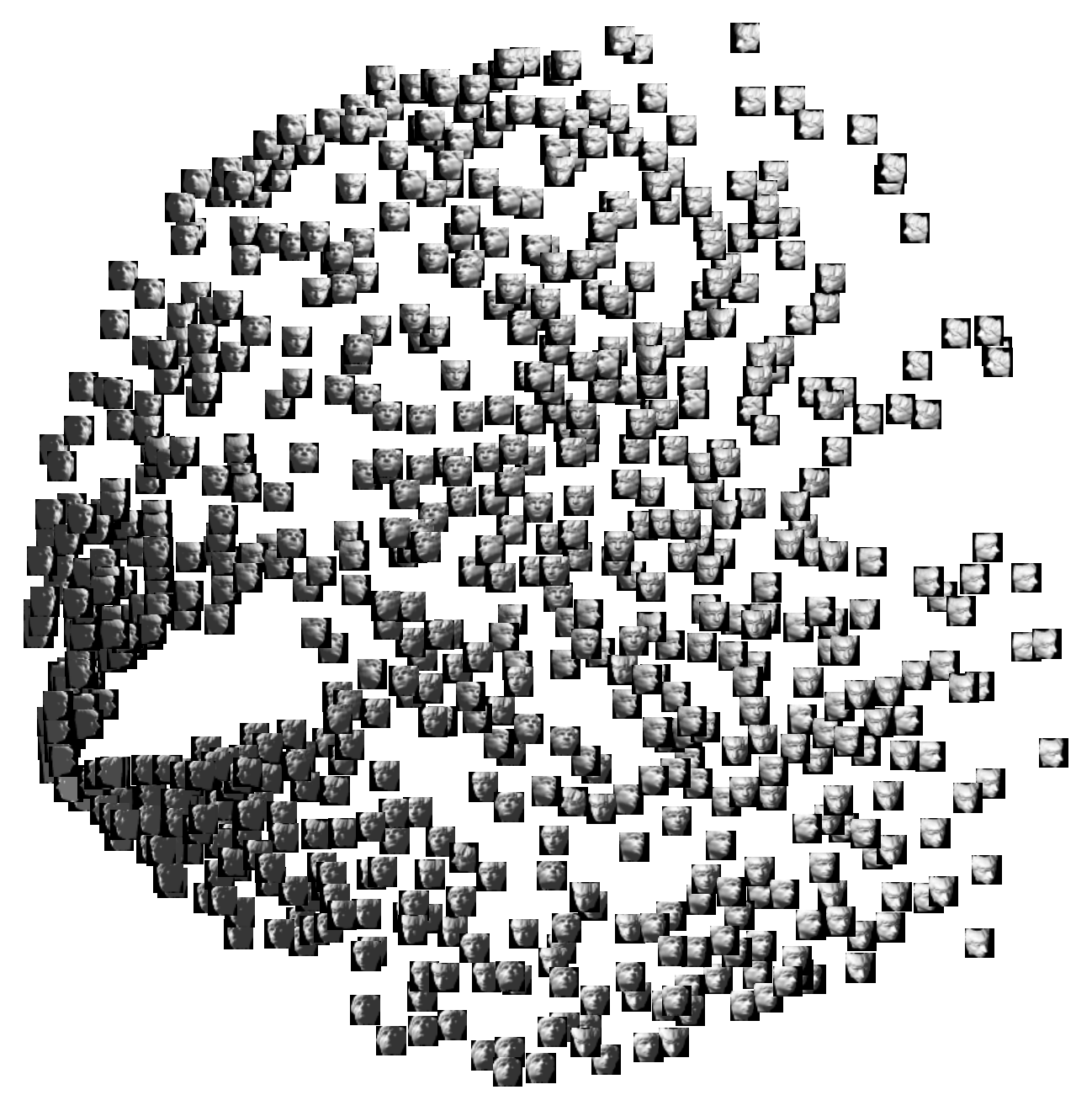}
        \caption{GW-MDS}
    \end{subfigure}
    \caption{Qualitative analysis of dimensionality reduction algorithms. Better seen on screen. While \gls{pca} and \gls{ewca} are linear, \gls{mds} and \gls{gw}-\gls{mds} (ours) are non-linear \gls{dr} strategies. In (a -- d), we show the low-dimensional representations of MNIST, whereas (e -- h) shows the representations for the faces dataset of~\cite{tenenbaum2000global}.}
    \label{fig:quali_analysis}
\end{figure*}
From a practical perspective, we leverage previous results in \gls{ot} for minimizing equation~(\ref{eq:gw_mds}), as it involves a nested minimization problem, with respect $Y_{it} = (y_{1}^{(it)},\cdots,y_{n}^{(it)})$, and $\pi \in \Pi(\hat{\mu},\hat{\nu})$. We do so, by alternating the minimization with respect to these variables. In a nutshell, at iteration $it$ and for a fixed $Y_{it}$, we solve for $\pi_{it}$ using equation~(\ref{eq:gromov-wasserstein}). Then, for a fixed $\pi_{it}$, we update $Y_{it+1}$ using \gls{gd} according to,
\begin{align*}
    Y_{it+1} = Y_{it} - \eta \nabla_{Y} GW(\hat{\mu},\hat{\nu}_{it}),
\end{align*}
where $\hat{\nu}_{it}$ is the empirical measure with support $Y_{it}$. This strategy is theoretically justified via~\cite{afriat1971theory}. The practical implementation of our algorithm is done in Pytorch~\cite{paszke2019pytorch} for automatic differentiation and Python Optimal Transport~\cite{flamary2021pot} for \gls{ot} related routines. Our code will be released upon acceptance. We summarize our proposal in Algorithm~\ref{alg:gw_mds}.

\begin{algorithm}
	\caption{GW-MDS} 
 \textbf{Input:} Data points $X = (x_{1}, \cdots, x_{n})$, $x_{i} \in \mathbb{R}^{p}$\\
 \textbf{Result:} Representations $Y = (y_{1}, \cdots, y_{n})$, $y_{i} \in \mathbb{R}^{d}$
        \begin{algorithmic}[1]
             \State Initialize $Y_{0} = (y_{1}^{(0)}, \cdots, y_{n}^{(0)})$
             \State $(C_{X})_{i,k} \leftarrow d_{\mathcal{X}}(x_{i},x_{k})$.
		      \For {$it=1,2,\ldots,N_{it}$}
                \State $\hat{\nu}_{it} \leftarrow n^{-1}\sum_{j=1}^{n}\delta_{y_{j}^{(it)}}$
			    \State $\pi_{it} \leftarrow \text{OT-GW}(\hat{\mu}, \hat{\nu}_{it})$ \Comment{using eq. 4.}
                \State $Y_{it+1} \leftarrow Y_{it} - \eta \nabla_{Y}GW(\hat{\mu},\hat{\nu}_{it})$
			\EndFor
	    \State Alignment $Y^{\star}_{i} \leftarrow n\sum_{j=1}^{n}\pi_{ij}^{\star}Y^{\star}_{j}$
        \end{algorithmic} 
    \label{alg:gw_mds}
\end{algorithm}

\vspace{1mm}
\noindent\textbf{Representation Initialization.} From the point of view of the minimization in equation~(\ref{eq:gw_mds}), one needs to determine an initialization for $y_{1},\cdots,y_{n}$. We propose two strategies. First, one may draw $y_{i} \sim \mathcal{N}(\mathbf{0}_{d},\mathbf{I}_{d})$ at random. Second, one may perform some \gls{dr} prior to our algorithm, such as \gls{pca}, so that $y_{i} = Wx_{i}$, where $W \in \mathbb{R}^{d \times p}$.

\vspace{1mm}
\noindent\textbf{Representation Alignment.} A major feature in our algorithm, is that we define a new notion of stress based on the \gls{ot} plan $\pi^{\star}$, as given in equation~(\ref{eq:gromov_wasserstein}). In this sense, one loses a direct correspondence between $x_{i}$ and $y_{i}$. However, it is possible to use $\pi^{\star}$ to align high-dimensional points and their representation, via the mapping $Y \mapsto n\sum_{j=1}^{n}\pi_{ij}^{\star}Y^{\star}_{j}$. In visualization tasks such that the order of points is important (e.g., manifold learning). This step can be done after \gls{gd} stops.

\vspace{1mm}

\noindent\textbf{Computational Complexity.} The complexity of Algorithm~\ref{alg:gw_mds} is dominated by the calculation of the transport plan $\pi_{it}$, which involves solving a \gls{gw} problem, which has $\mathcal{O}(n^{3})$ per iteration. As we demonstrate in our experiments, this is not prohibitive. We leave the question of improving the complexity for future works.

\section{Experiments and Discussion}\label{sec:experiments}

In this section, we apply our method to toy examples in manifold learning, as well as real world datasets. Our main point of comparison is with \gls{mds}, but we also consider other \gls{ot}-based dimensionality reduction algorithms, such as \gls{ewca}~\cite{collas2023entropic}.

\begin{figure}[ht]
    \centering
    \begin{subfigure}{0.3\linewidth}
        \includegraphics[width=\linewidth]{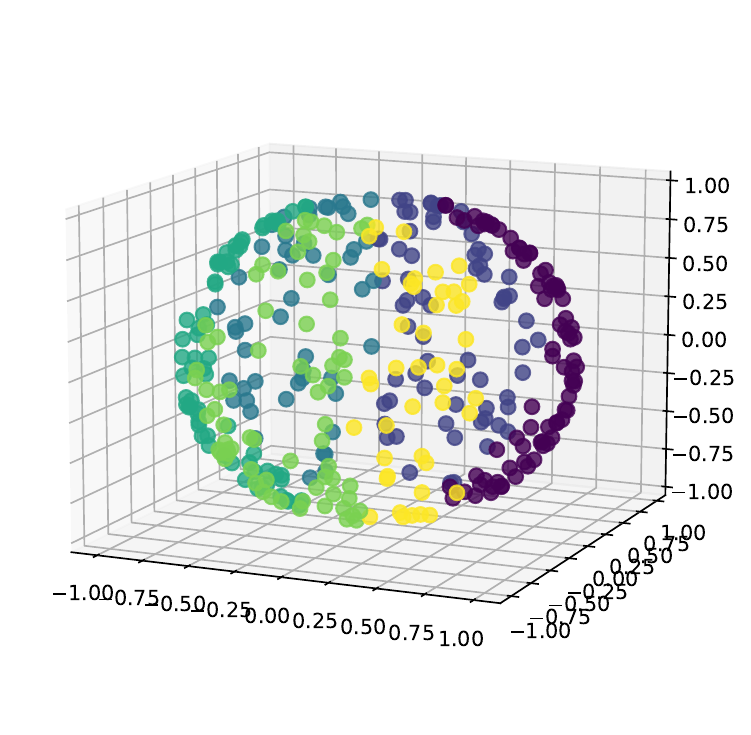}
        \caption{Manifold}
    \end{subfigure}\hfill
    \begin{subfigure}{0.3\linewidth}
        \includegraphics[width=\linewidth]{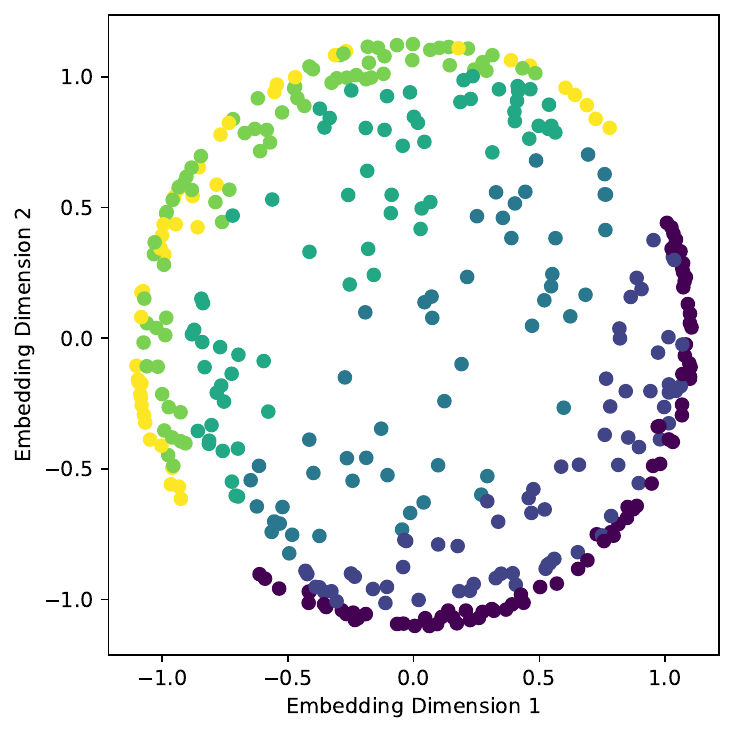}
        \caption{MDS}
    \end{subfigure}\hfill
    \begin{subfigure}{0.3\linewidth}
        \includegraphics[width=\linewidth]{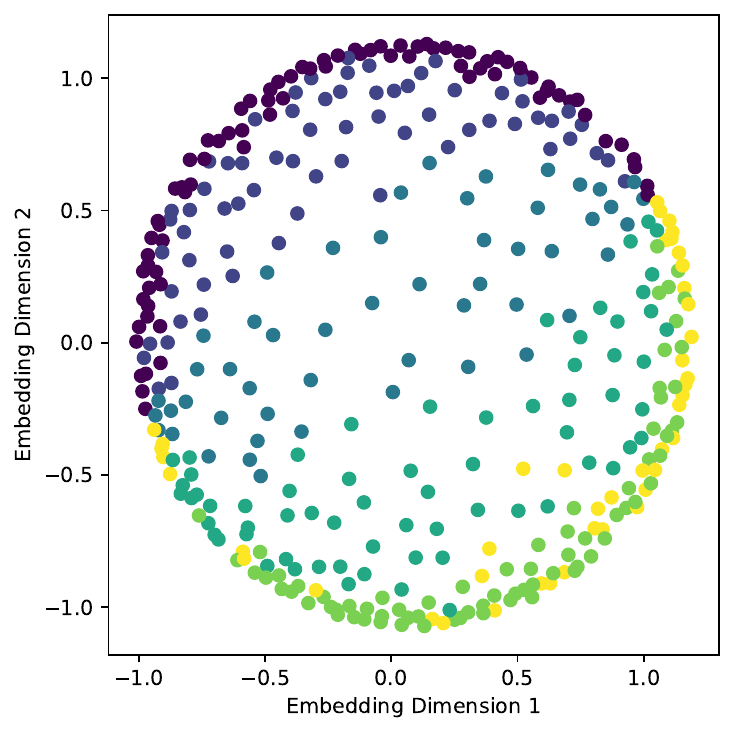}
        \caption{GW-MDS}
    \end{subfigure}\\
    \begin{subfigure}{0.3\linewidth}
        \includegraphics[width=\linewidth]{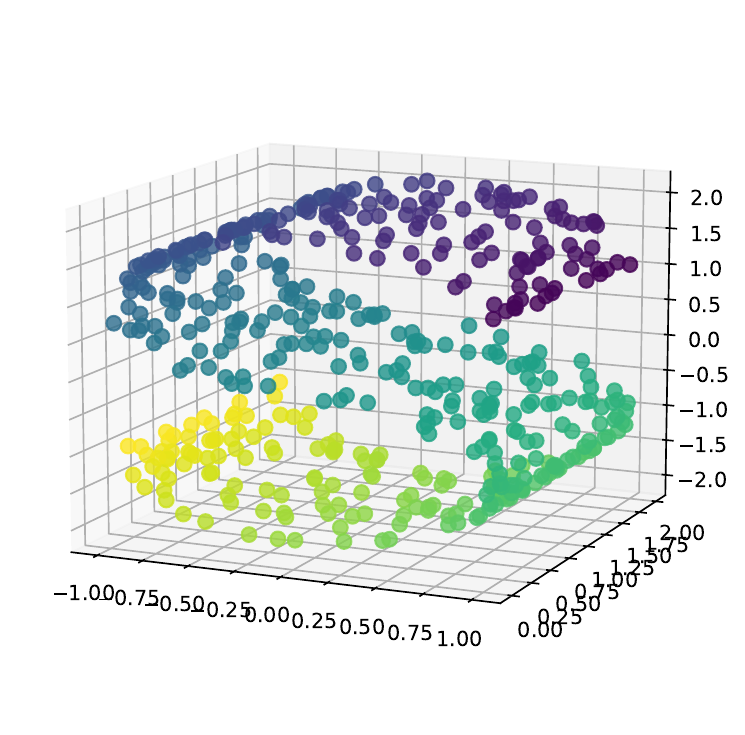}
        \caption{Manifold}
    \end{subfigure}\hfill
    \begin{subfigure}{0.3\linewidth}
        \includegraphics[width=\linewidth]{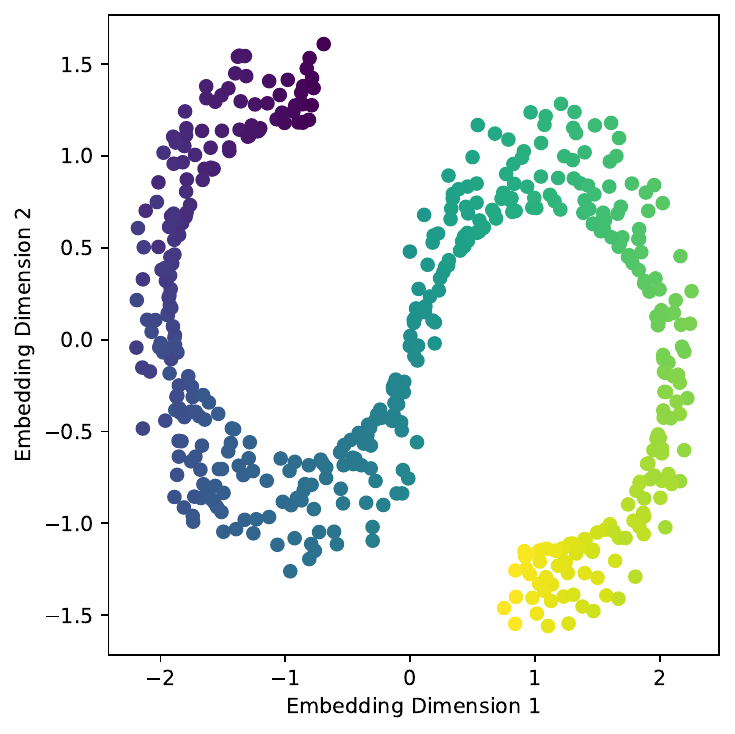}
        \caption{MDS}
    \end{subfigure}\hfill
    \begin{subfigure}{0.3\linewidth}
        \includegraphics[width=\linewidth]{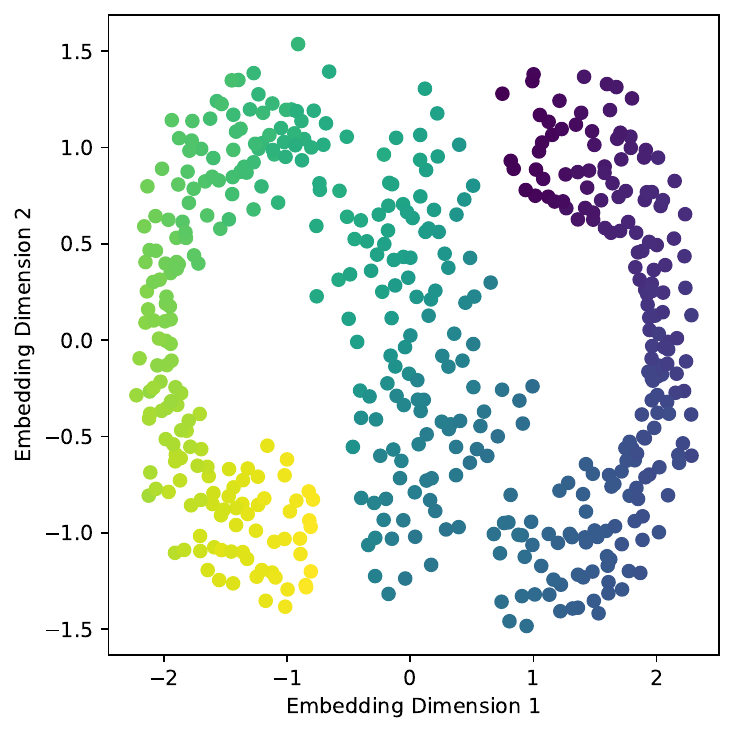}
        \caption{GW-MDS}
    \end{subfigure}
    \caption{Panels (a) and (d) show the toy manifolds, panels (b) and (e) show the embeddings generated by MDS, while panels (c) and (f) describe the results of GW-MDS. The figure illustrates how GW-MDS better preserves structural relationships and distances in low-dimensional space compared to MDS, highlighting its effectiveness in maintaining the original data topology}
    \label{fig:manif}
\end{figure}

\vspace{1mm}
\begin{figure}[ht]
    \centering
    \begin{subfigure}{0.48\linewidth}
        \centering
        \includegraphics[width=\linewidth]{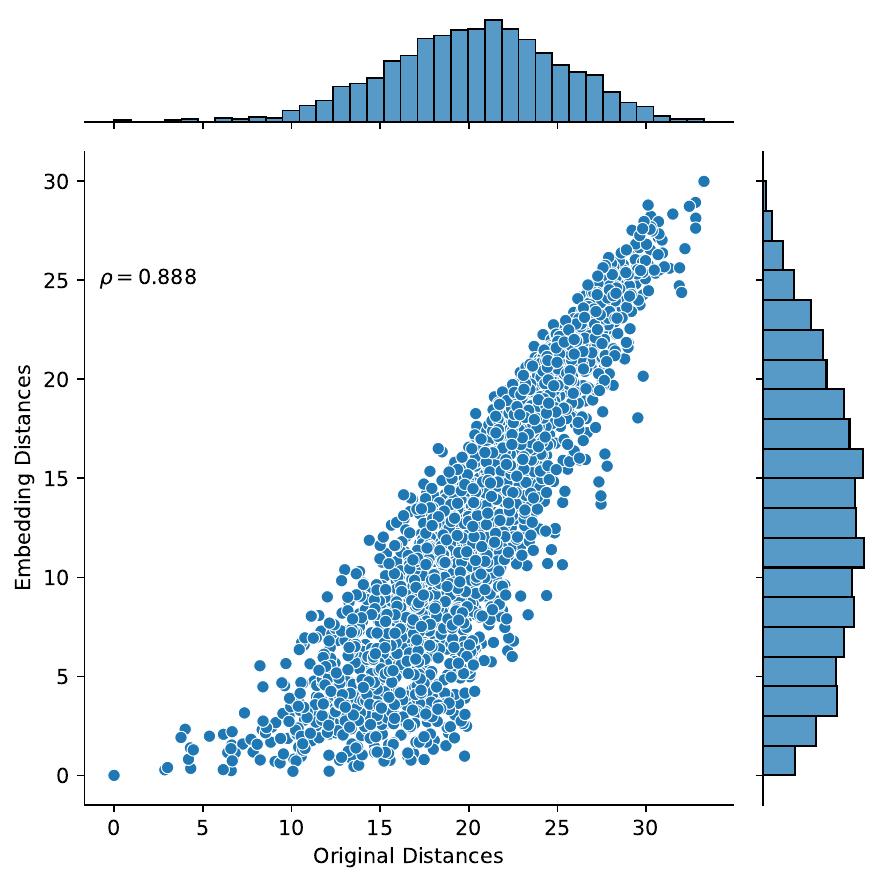}
        \caption{PCA}
    \end{subfigure}\hfill
    \begin{subfigure}{0.48\linewidth}
        \centering
        \includegraphics[width=\linewidth]{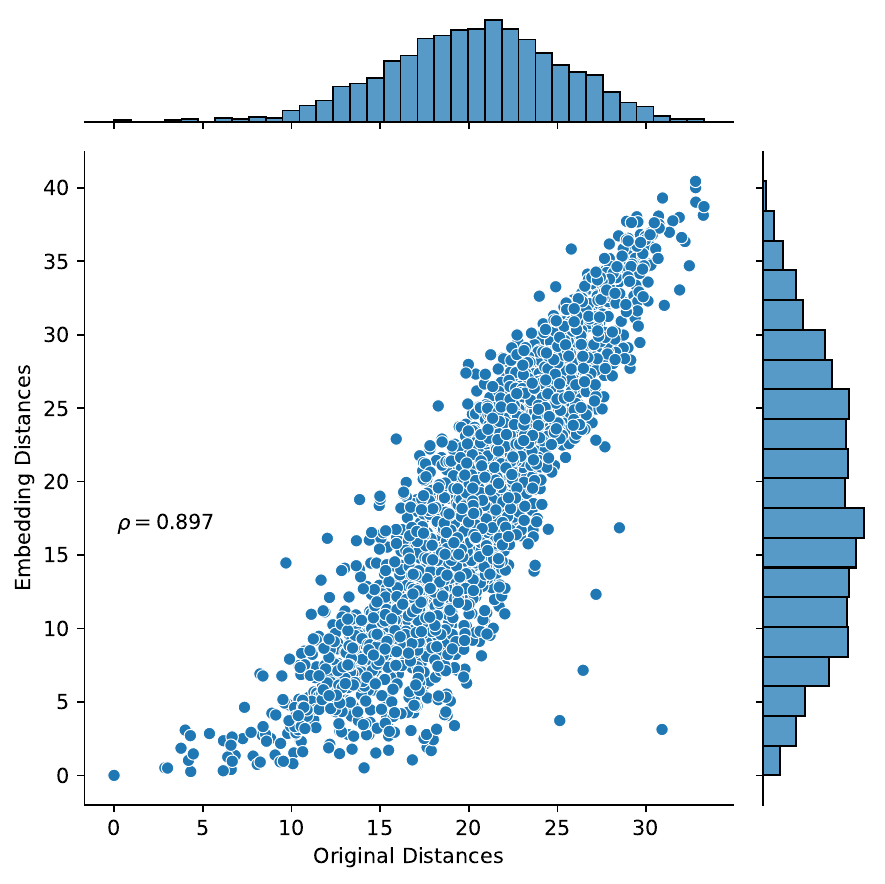}
        \caption{MDS}
    \end{subfigure}\hfill
    \begin{subfigure}{0.48\linewidth}
        \centering
        \includegraphics[width=\linewidth]{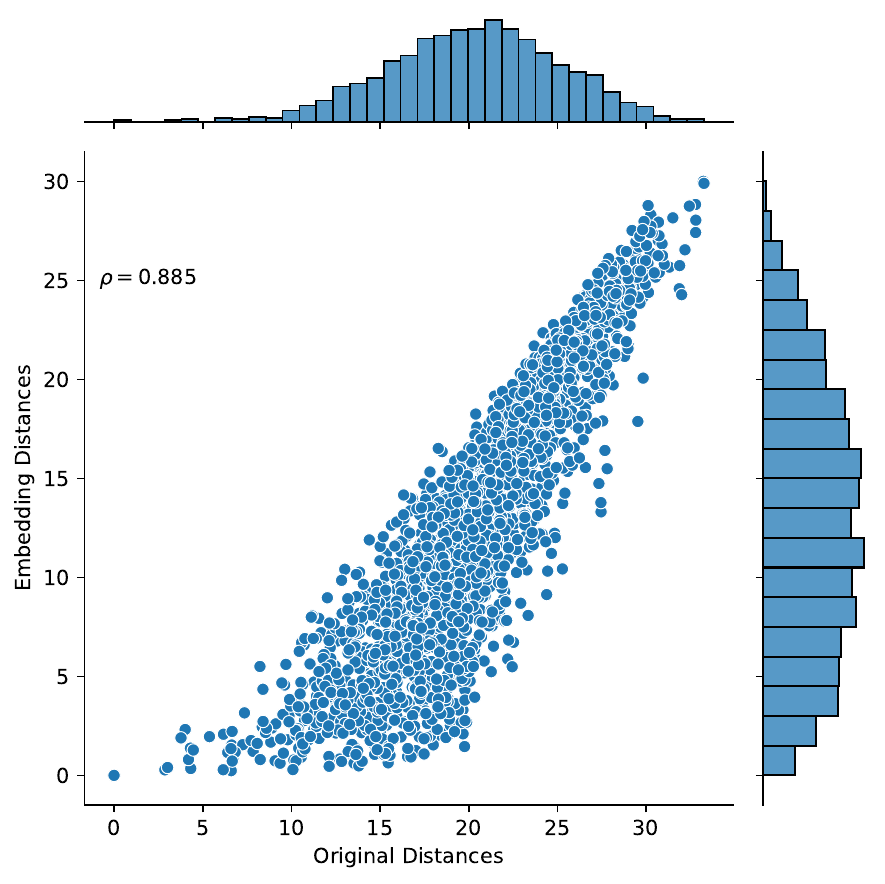}
        \caption{EWCA}
    \end{subfigure}\hfill
    \begin{subfigure}{0.48\linewidth}
        \centering
        \includegraphics[width=\linewidth]{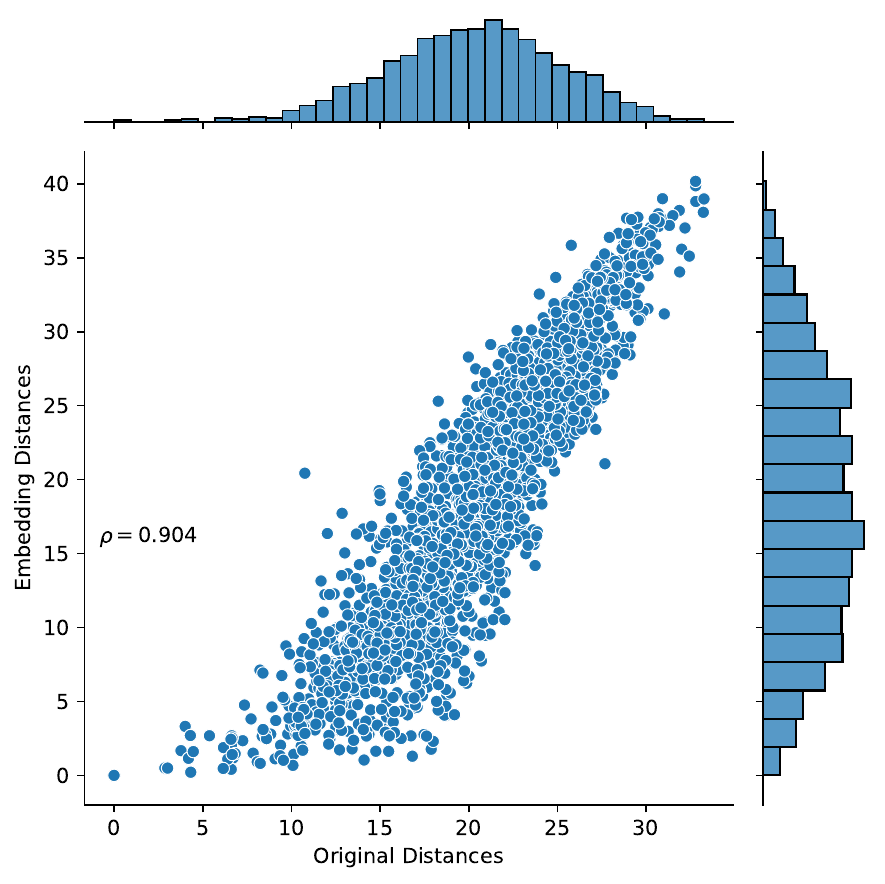}
        \caption{GW-MDS}
    \end{subfigure}
    \caption{Scatter plot of pairwise distances in $\mathcal{X}$ and $\mathcal{Y}$. Distances are computed between points in the faces dataset of~\cite{tenenbaum2000global}. Overall, \gls{gw}-\gls{mds} yields embeddings that better preserve the pairwise distances in $\mathcal{X}$.}
    \label{fig:jointplot_mnist_2}
\end{figure}
\vspace{1mm}
\noindent\textbf{Experiments on toy datasets.} We apply the GW-MDS algorithm to synthetic datasets commonly used in manifold learning to demonstrate the effectiveness of our method in preserving the intrinsic structure of high dimensional data is shown in Fig~\ref{fig:manif}.

Before applying the \gls{dr} techniques, the datasets were pre-processed by normalizing each feature to ensure consistent scaling and to improve the performance of the algorithm. The toy datasets provide a controlled environment where the underlying manifold structure is known, allowing for a clear comparison between different \gls{dr} techniques. Our experiments show that GW-MDS consistently outperforms traditional methods like MDS in maintaining the distances between data points, which is critical for applications where the preservation of the original data topology is essential. Additionally, the results highlight the robustness of GW-MDS, especially in scenarios involving complex, non-linear data structures, showcasing its potential for broader applications in machine learning and data analysis.

\vspace{1mm}

\noindent\textbf{Experiments on realistic datasets.} In this part, we experiment with the \gls{dr} of high dimensional realistic datasets. Especially, we use MNIST~\cite{lecun1998mnist}, Faces~\cite{tenenbaum2000global}. The MNIST dataset was chosen due to its use in dimensionality reduction work with Gromov-Wasserstein ~\cite{van2024distributional} and ~\cite{clark2024generalized}. These datasets consist of gray-scale images of shape $(28, 28)$ and $(64, 64)$ respectively. As pre-processing steps, we convert each image to 32-bit float encoding, then normalize each pixel by its maximum value, i.e., $255$. The images are then flattened into vectors. This creates datasets with an increasing number of dimensions, that is, $784$, $4,096$, respectively. A qualitative comparison between the embeddings obtained by \gls{pca}, \gls{mds}, \gls{ewca} and \gls{gw}-\gls{mds} is shown in Fig.~\ref{fig:quali_analysis}.

Furthermore, we quantify how well the compared algorithms capture the high-dimensional distances through their embeddings. We do so through two metrics, namely, the stress introduced in equation~(\ref{eq:mds}), and the Pearson correlation coefficient between distances in the ambient space, $d_{\mathcal{X}}$, and distances in the embedding space, $d_{\mathcal{Y}}$, that is,
\begin{align}
    \rho = \dfrac{\text{cov}(d_{\mathcal{X}}, d_{\mathcal{Y}})}{\sigma(d_{\mathcal{X}})\sigma(d_{\mathcal{Y}})},\label{eq:corrcoef}
\end{align}
where $\text{cov}(X, Y)$ and $\sigma(X)$ is the covariance and standard deviation for random variables $X$ and $Y$. We summarize our quantitative analysis in Table~\ref{tab:quant_analysisI}.

\begin{table}[!h]
\centering
\caption{Quantitative analysis of dimensionality reduction algorithms using Pearson's correlation coefficient (Equation~\eqref{eq:corrcoef}).}
\vspace{2mm}
\label{tab:quant_analysisI}
\begin{tabular}{|l|c|c|c|c|c|c|c|}
\hline
\textbf{Method} & \textbf{MNIST} & \textbf{Faces} & \textbf{Sphere} & \textbf{S-Curve} & \textbf{Torus}& \textbf{Mobius}\\
\hline
MDS    & 0.643          & 0.897          & 0.781           & 0.949   & \textbf{0.993} & \textbf{0.952}  \\
GW-MDS & \textbf{0.646} & \textbf{0.904} & \textbf{0.826}  & 0.966  & \textbf{0.993} & 0.947\\
PCA    & 0.523          & 0.888          & 0.817           & \textbf{0.970} & \textbf{0.993} & 0.934\\
EWCA   & 0.526          & 0.883          & 0.819           & \textbf{0.970} & \textbf{0.993} & 0.73\\
\hline
\end{tabular}
\end{table}
To give a global view on the distribution of distances in $\mathcal{X}$, and $\mathcal{Y}$, we show, in Fig.~\ref{fig:jointplot_mnist_2}, a scatter plot between these two variables in the context of MNIST. In general, the distances are correlated, indicating that all algorithms capture the geometry of high-dimensional data. However, non-linear \gls{dr} algorithm such as \gls{mds} and \gls{gw}-\gls{mds} still have an advantage over linear ones, as these can capture more complex relationships.

\begin{table*}[!h]
\centering
\caption{Quantitative analysis of dimensionality reduction algorithms using Pearson's correlation coefficient (Equation~\eqref{eq:corrcoef}). With the version of the algorithm using geodesic distance.}
\vspace{2mm}
\label{tab:quant_analysisII}
\begin{tabular}{|l|c|c|c|c|c|c|c|c|}
\hline
\textbf{Method} & \textbf{MNIST} & \textbf{Faces} & \textbf{Swiss roll} & \textbf{S-Curve}  & \textbf{Torus} & \textbf{Mobius} & \textbf{Sphere} \\
\hline
GW-MDS  & \textbf{0.7887} & 0.9306        & \textbf{0.9993} & \textbf{0.9993}  & \textbf{0.9719} & \textbf{0.9499} & 0.9623\\
Isomap  & 0.7697          & \textbf{0.9561} & 0.9986          & 0.9988     & 0.9698  & 0.9441 & \textbf{0.9695} \\
\hline
\end{tabular}
\end{table*}
Building on these observations of correlated distances, especially when non-linear relationships are likely, we can further refine the approach by incorporating geodesic distances into \( d_{\mathcal{X}}(x_i, x_j) \) in equation~\ref{eq:gromov_wasserstein}. This variation of GW-MDS becomes particularly useful when the data is presumed to lie on a manifold, allowing the algorithm to capture the ‘curved’ geometry of the dataset via nearest-neighbor graphs.

\begin{figure}[h]
    \centering
    \begin{subfigure}{0.33\linewidth}
        \includegraphics[width=\linewidth]{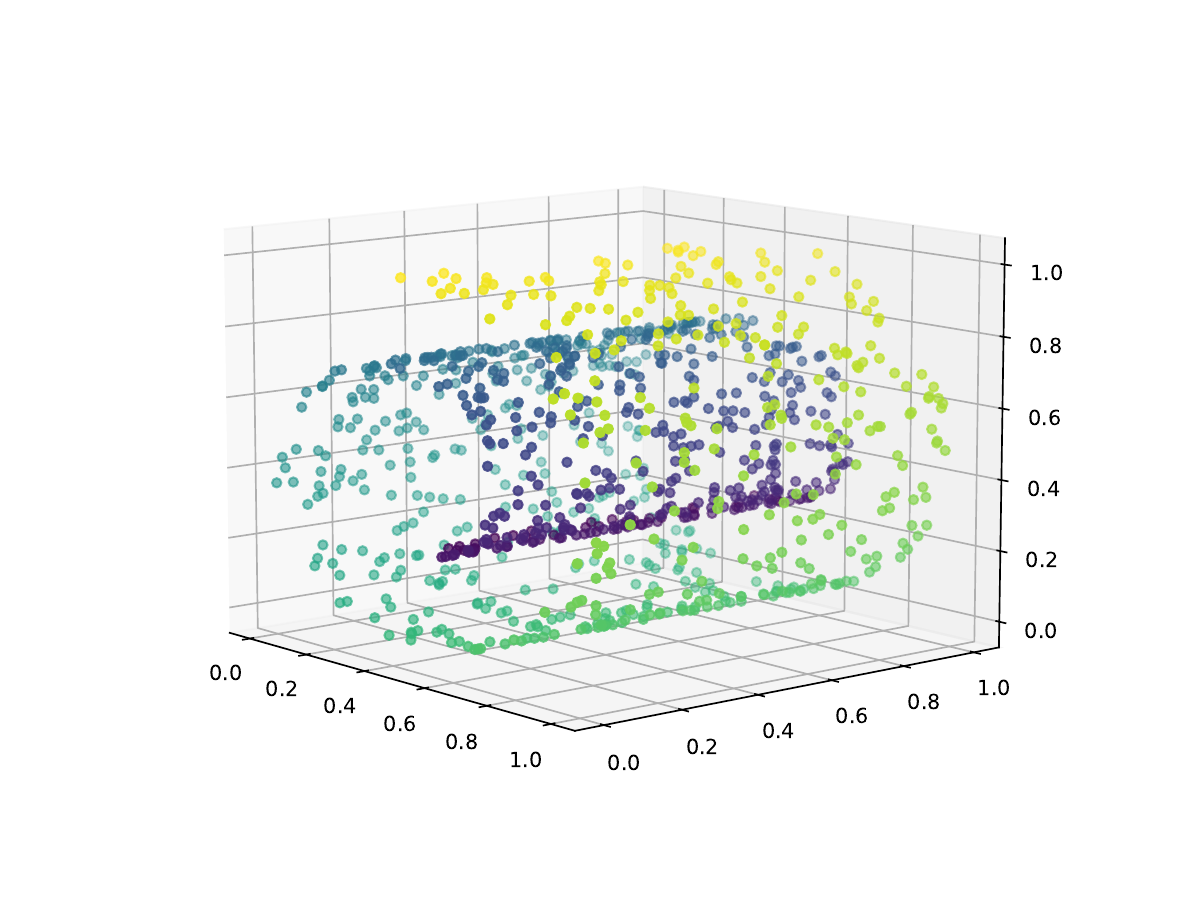}
        \caption{Manifold}
    \end{subfigure}\hfill
    \begin{subfigure}{0.33\linewidth}
        \includegraphics[width=\linewidth]{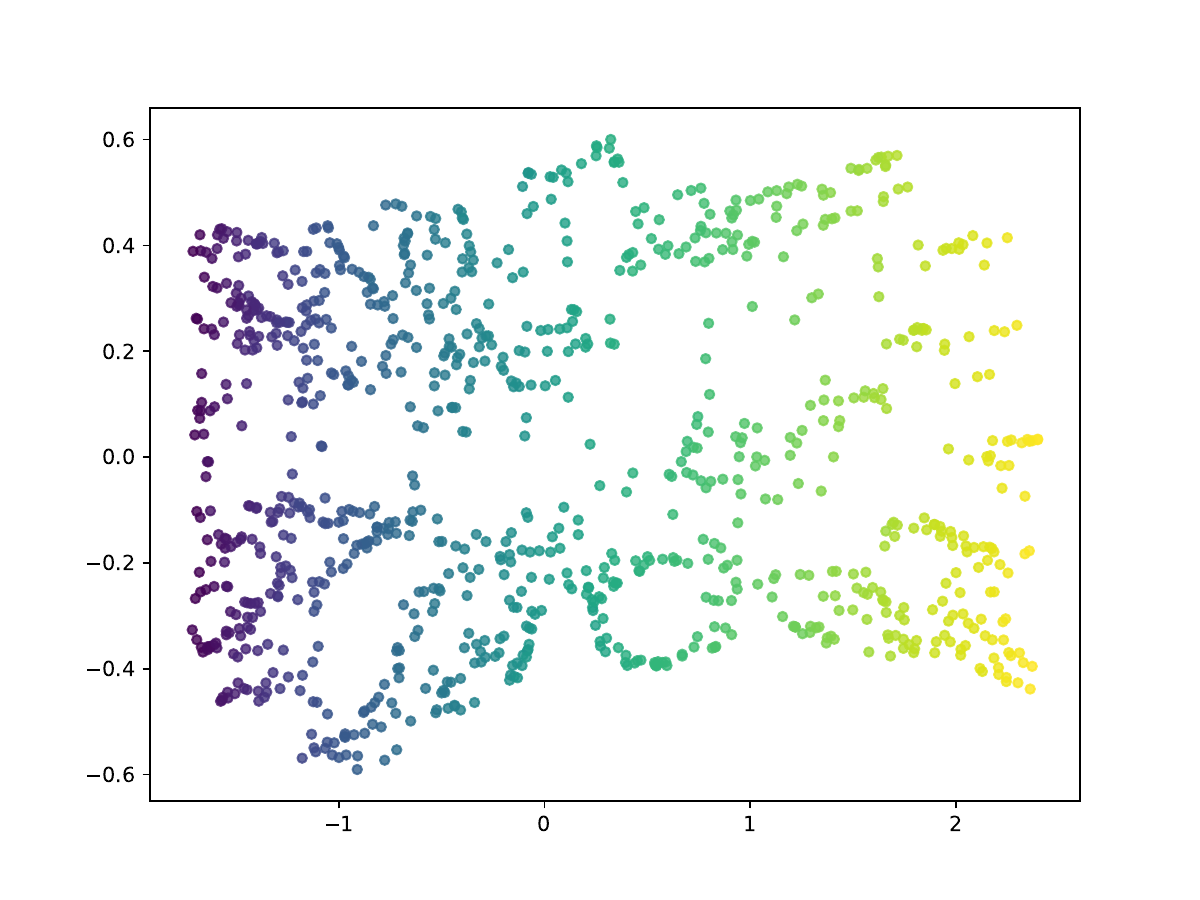}
        \caption{Isomap}
    \end{subfigure}\hfill
    \begin{subfigure}{0.33\linewidth}
        \includegraphics[width=\linewidth]{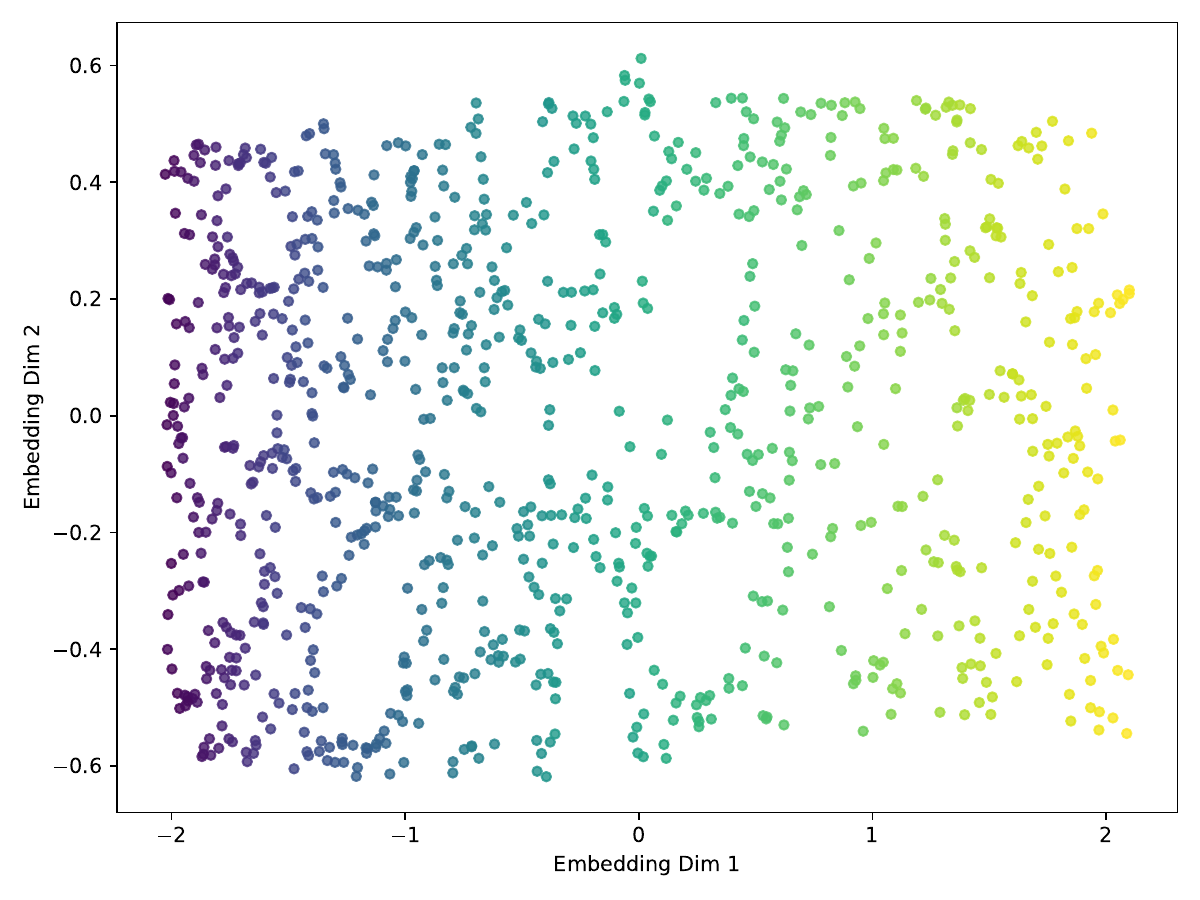}
        \caption{GW-MDS}
    \end{subfigure}\\
    \begin{subfigure}{0.33\linewidth}
        \includegraphics[width=\linewidth]{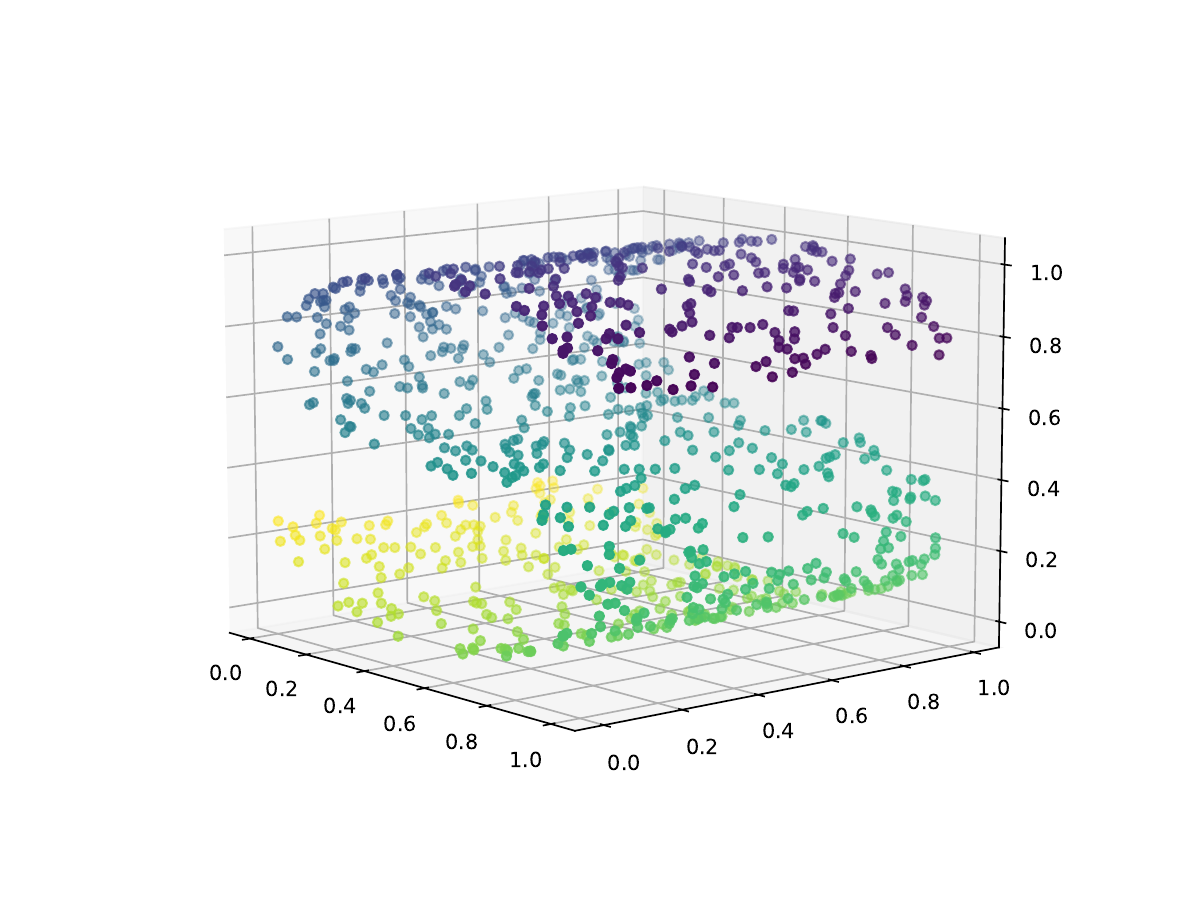}
        \caption{Manifold}
    \end{subfigure}\hfill
    \begin{subfigure}{0.33\linewidth}
        \includegraphics[width=\linewidth]{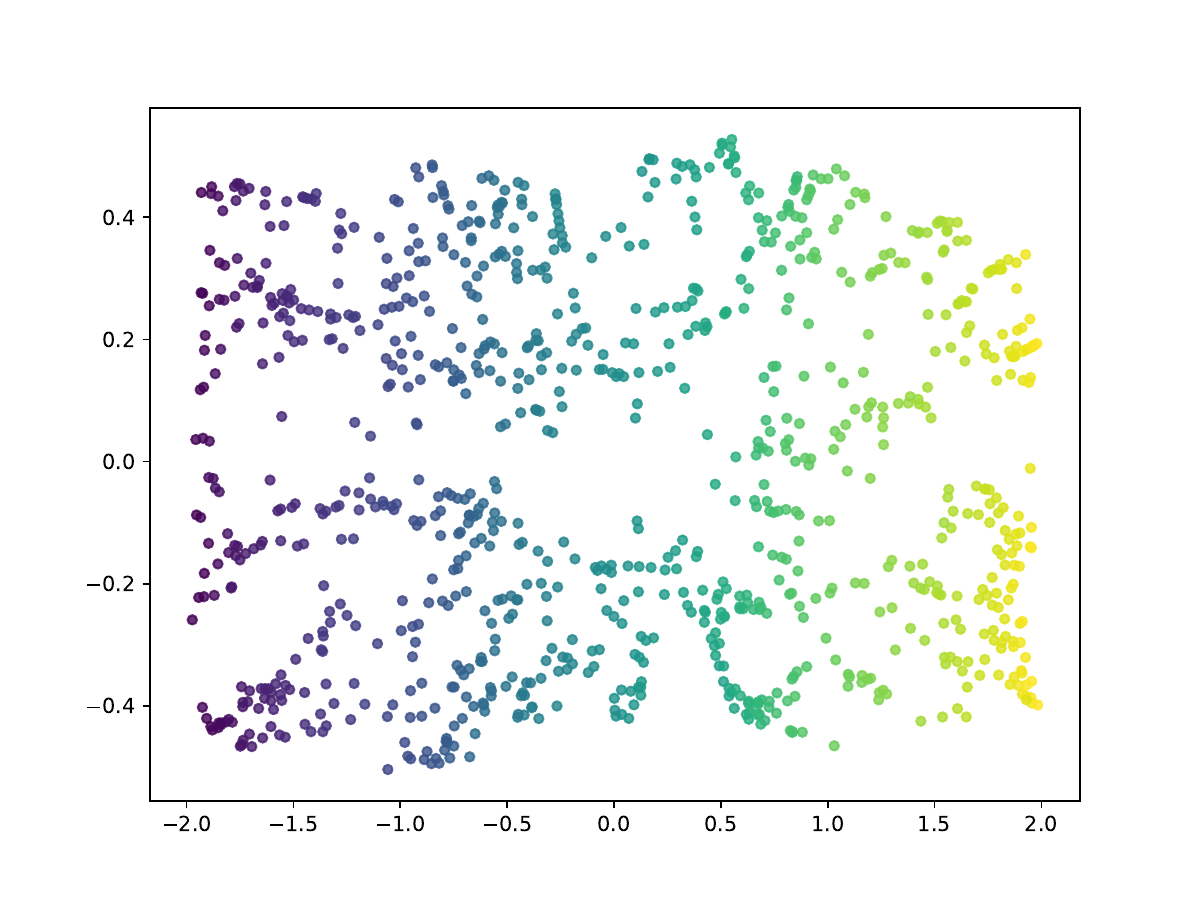}
        \caption{Isomap}
    \end{subfigure}\hfill
    \begin{subfigure}{0.33\linewidth}
        \includegraphics[width=\linewidth]{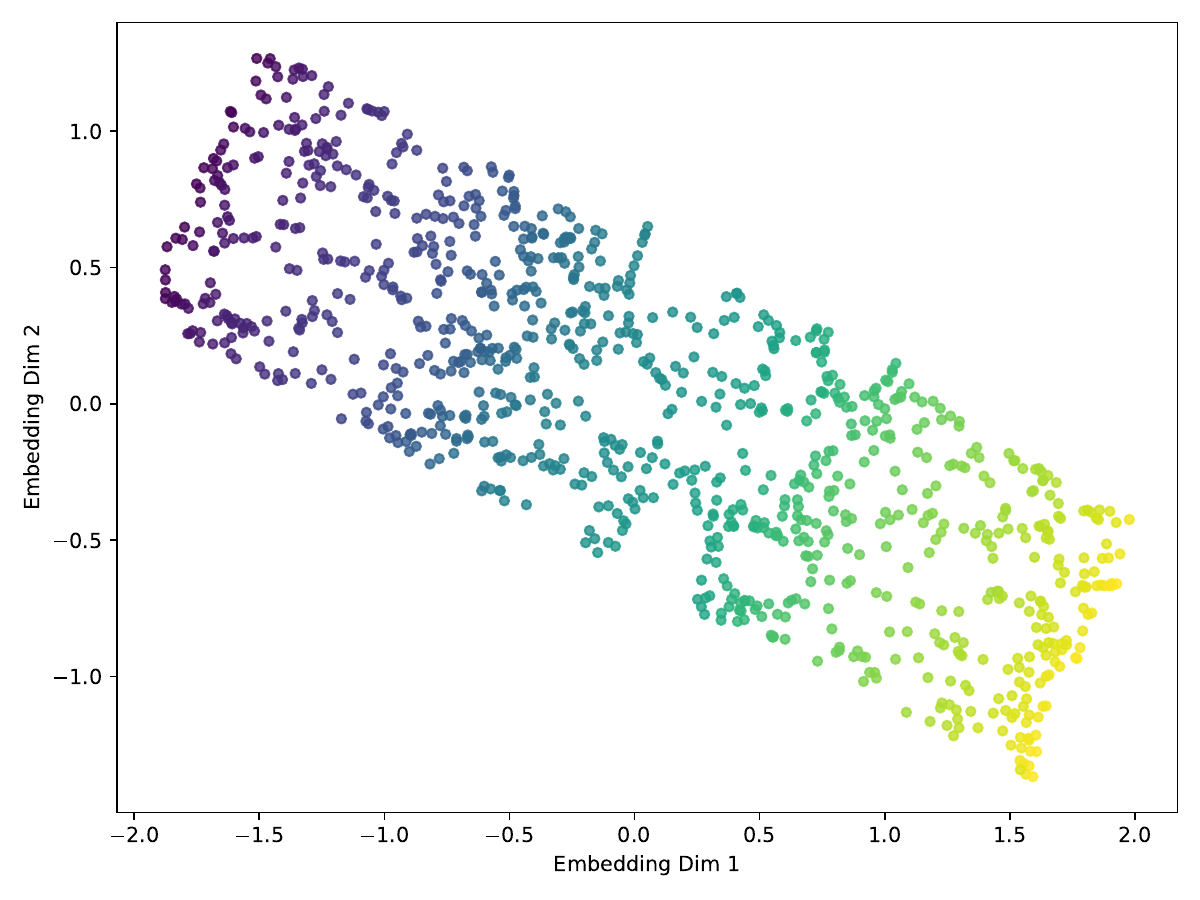}
        \caption{GW-MDS}
    \end{subfigure}
    \caption{Panels (a) and (d) show the toy manifolds, panels (b)
and (e) show the embeddings generated by Isomap, while the panels
(c) and (f) describe the results of the GW-MDS, using the geodesic distance.}
    \label{fig:manif2}
\end{figure}
\begin{figure}[h!]
    \centering
    \begin{subfigure}{0.5\linewidth}
        \includegraphics[width=\linewidth]{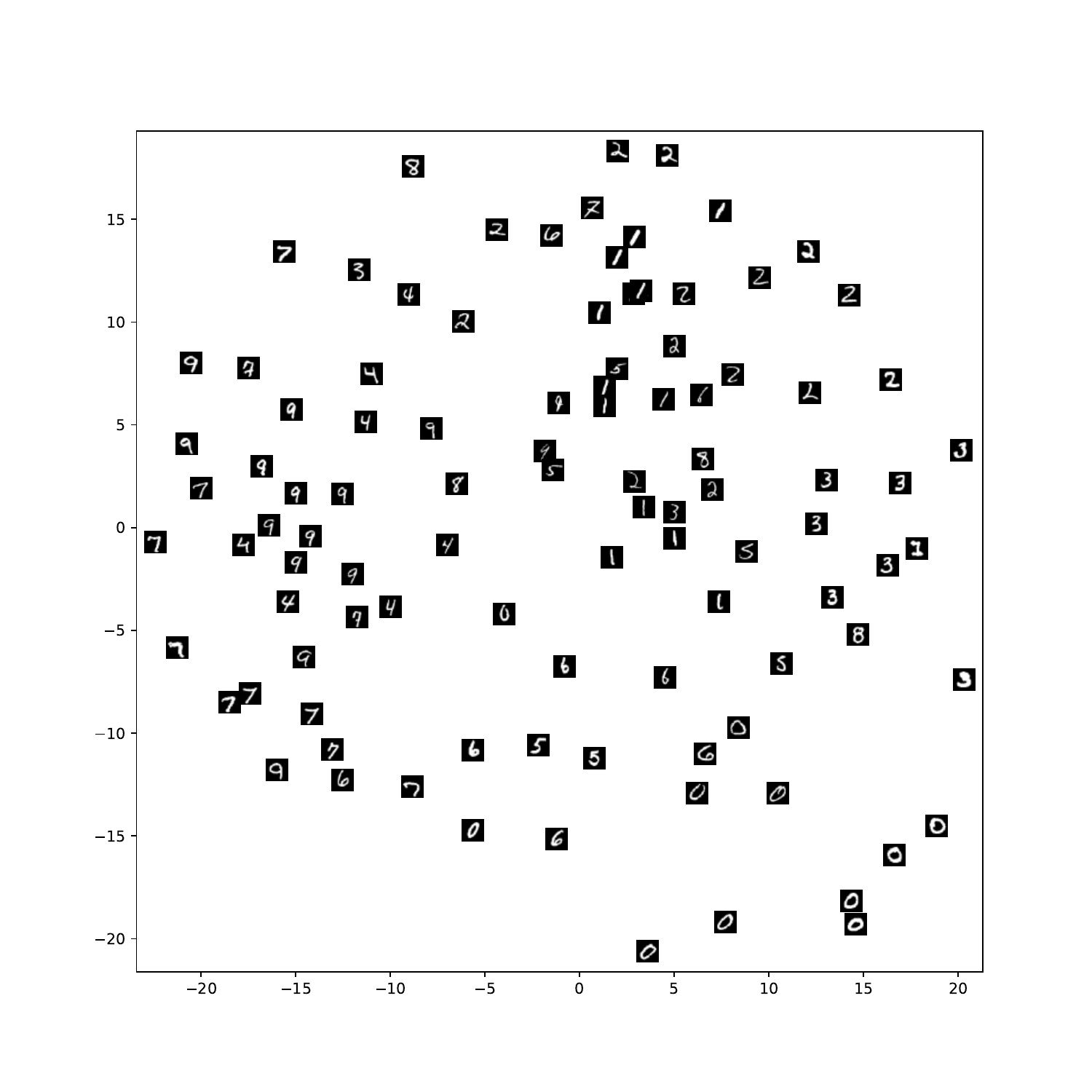}
        \caption{GW-MDS}
    \end{subfigure}\hfill
    \begin{subfigure}{0.5\linewidth}
        \includegraphics[width=\linewidth]{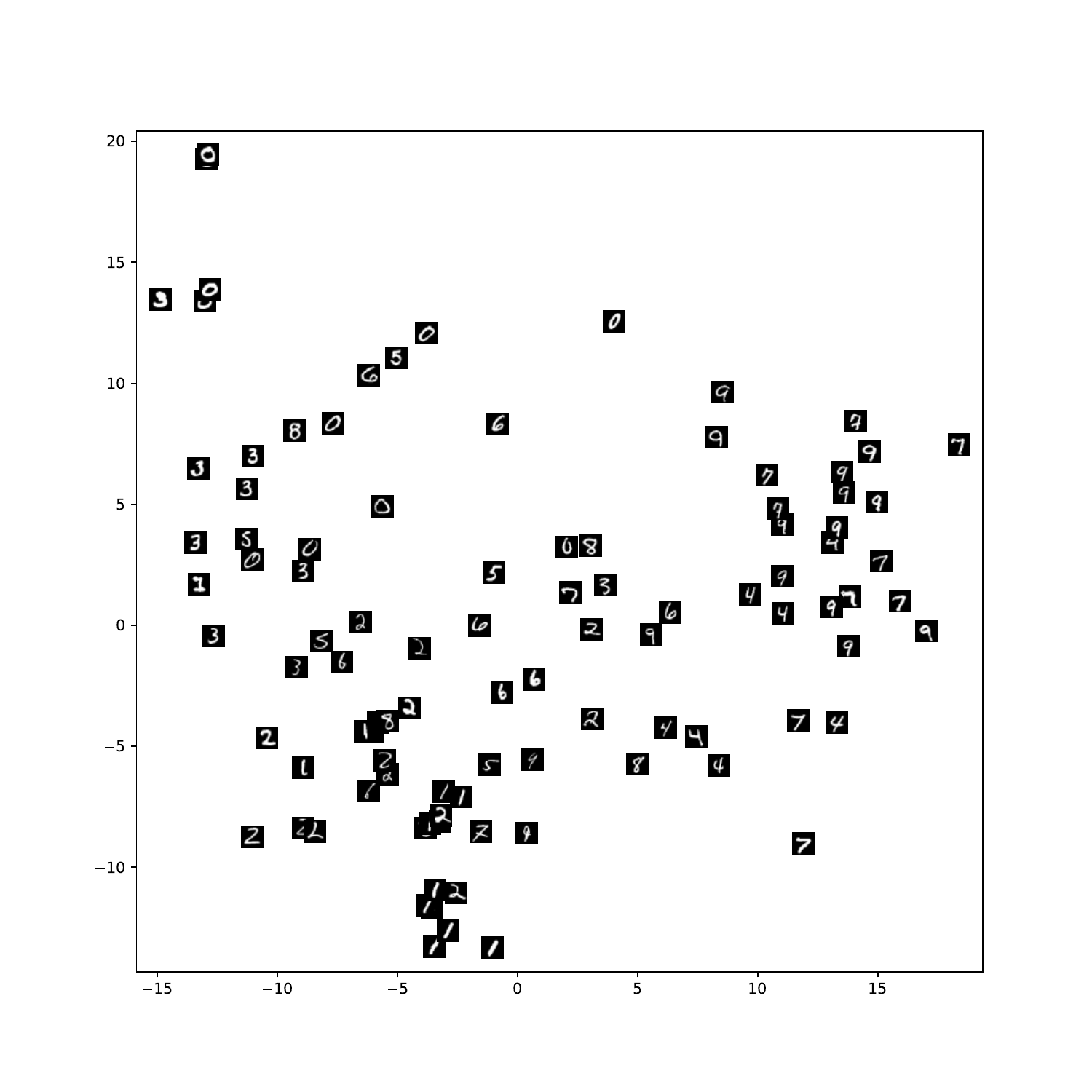}
        \caption{Isomap}
    \end{subfigure}
    \begin{subfigure}{0.5\linewidth}
        \includegraphics[width=\linewidth]{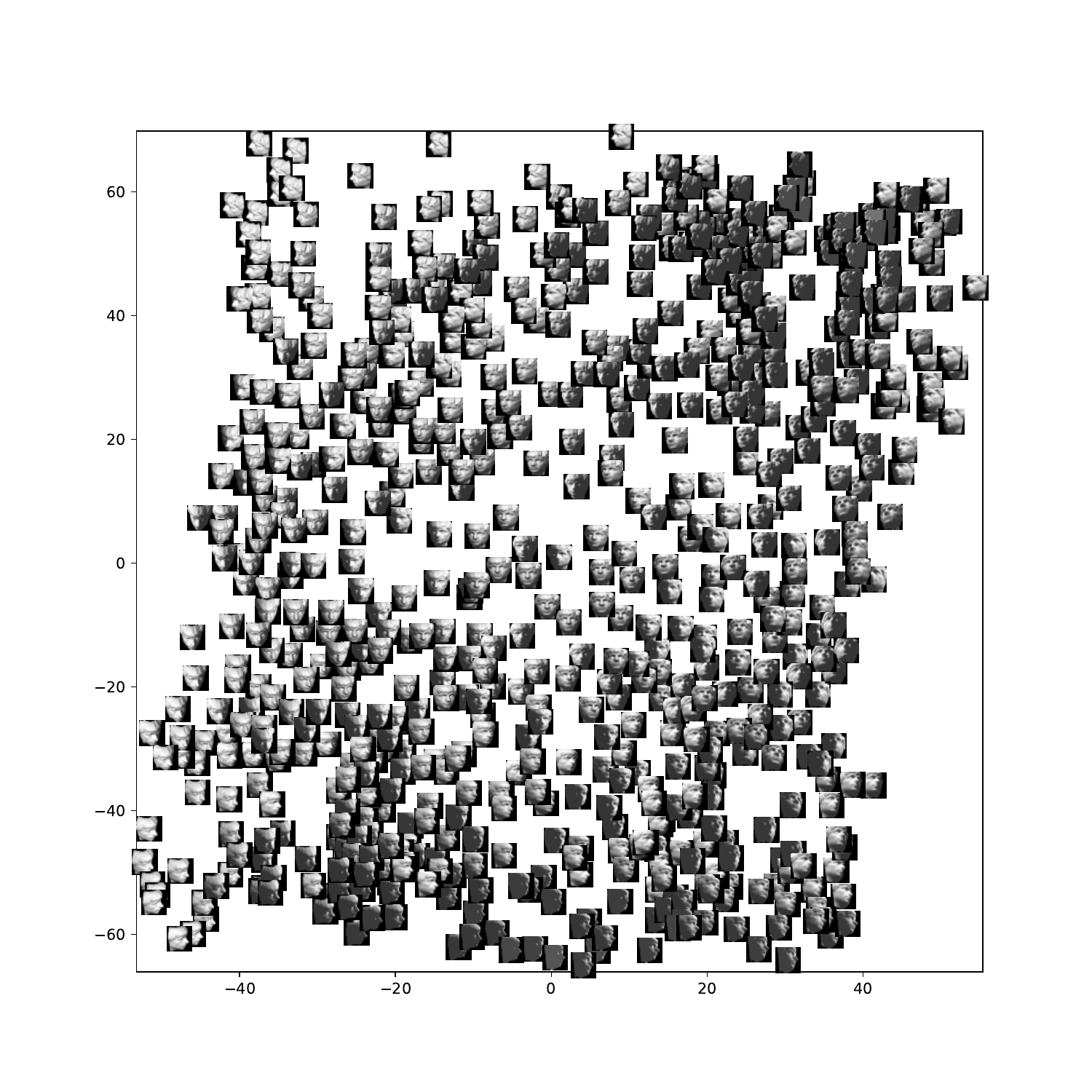}
        \caption{GW-MDS}
    \end{subfigure}\hfill
    \begin{subfigure}{0.5\linewidth}
        \includegraphics[width=\linewidth]{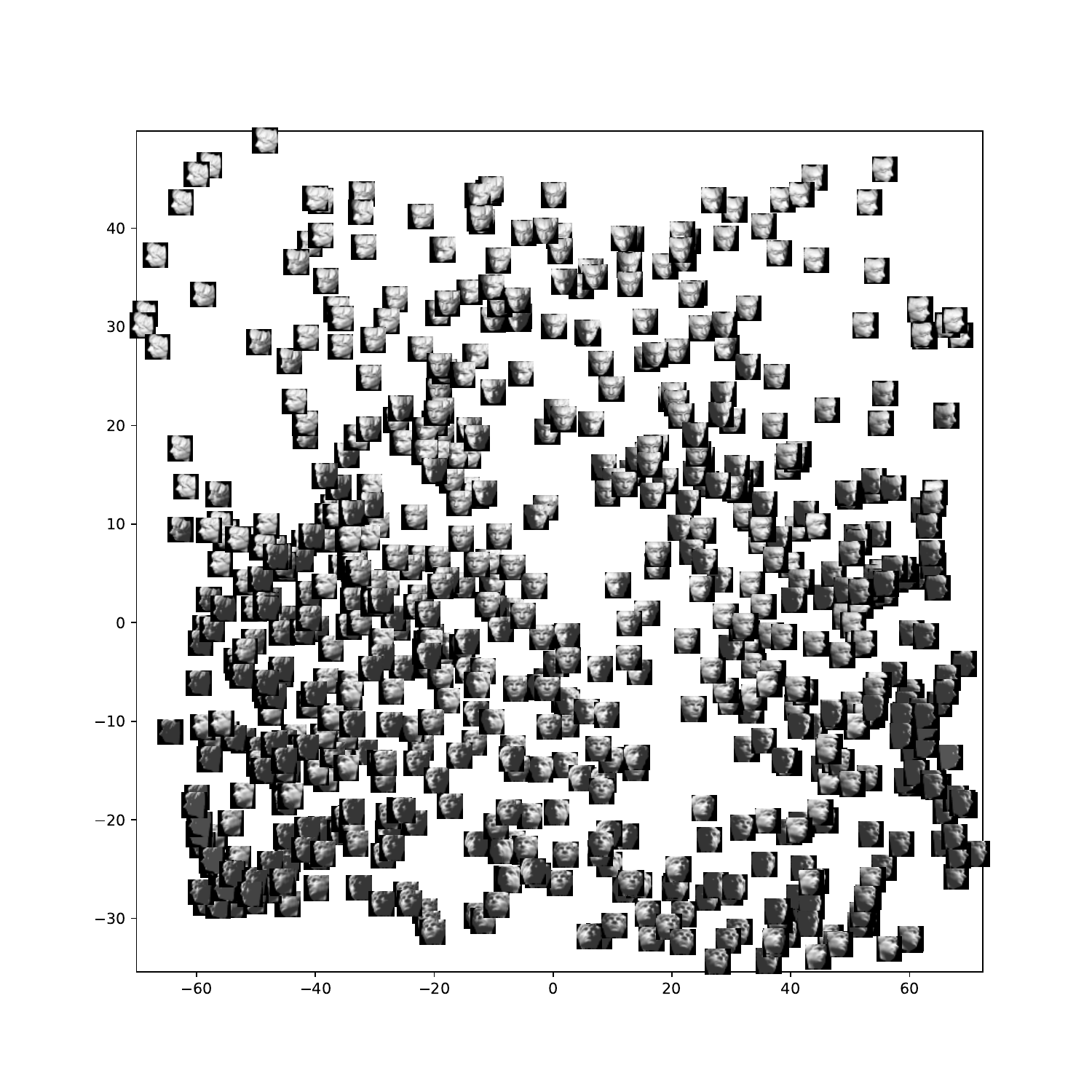}
        \caption{Isomap}
    \end{subfigure}
    \caption{(a) and (b) are the dimensionality reduction of the MNIST dataset, where (a) is the reduction using GW-MDS with the geodesic distance and (b) is the reduction using Isomap, whereas items (c) and (d), are the reduction using the Faces dataset, where, (c) is the reduction using GW-MDS with the geodesic distance and (d) using the Isomap algorithm.}
    \label{fig:fig:quali_analysis2}
\end{figure}

By introducing the geodesic distance in place of the Euclidean metric \(d_{\mathcal{X}}(x_i,x_j)\) in equation~\ref{eq:gromov_wasserstein}, we obtain a variant of the GW-MDS algorithm particularly well-suited for datasets that lie (ou are suspected to lie) on a manifold. In practical terms, this involves constructing a nearest-neighbor graph of the original data and estimating the pairwise distances along the edges of this graph, capturing the intrinsic geometry or “geodesy” of the data rather than merely its straight-line (Euclidean) distance. Consequently, this approach is especially powerful in settings where the data is highly non-linear or “curved” in its ambient space, as it more accurately reflects the local and global structure of such manifolds.

To illustrate the effectiveness of this variant, we performed tests on classic toy manifolds, as shown in Figure~\ref{fig:manif2}. These experiments provide a direct comparison between the proposed GW-MDS with geodesic distance and the well-known Isomap algorithm. The visual results demonstrate how our GW-MDS approach not only preserves local neighborhoods but also recovers the global manifold structure with high fidelity, often comparable or even superior to that of Isomap.

Moreover, we applied the same method to real-world datasets, specifically MNIST~\cite{lecun1998mnist} and Faces~\cite{tenenbaum2000global}, as depicted in Figure~\ref{fig:fig:quali_analysis2}. There, we present a side-by-side comparison of embeddings obtained via geodesic-based GW-MDS and those produced by Isomap. In addition to visually inspecting how well each approach unfolds the manifold in a low-dimensional space, we also computed the correlation of pairwise distances for each technique, summarized in Table~\ref{tab:quant_analysisII}. These quantitative results confirm that our geodesic-based GW-MDS can capture both local and global structures effectively, offering a compelling alternative for manifold learning tasks where conventional Euclidean distances may not suffice.

Table~\ref{tab:quant_analysisII}, we present the Pearson correlation coefficients for our geodesic-based GW-MDS algorithm and Isomap across four distinct datasets: MNIST, Faces, Swiss Roll, and S-Curve.  Overall, GW-MDS demonstrates strong performance, especially for the non-linear manifolds (Swiss Roll and S-Curve), where its correlations reach or exceed 0.9993. While Isomap also achieves competitive, scores particularly on the Faces dataset, GW-MDS tends to capture both local and global structures more effectively, reinforcing the idea that incorporating geodesic distances into the Gromov-Wasserstein framework can be advantageous for manifold learning tasks.

Having established the effectiveness of our geodesic-based approach, we next turn our attention to the optimization details.

\vspace{1mm}

\begin{figure}[h]
    \centering
    \begin{subfigure}{0.49\linewidth}
        \centering
        \includegraphics[width=\linewidth]{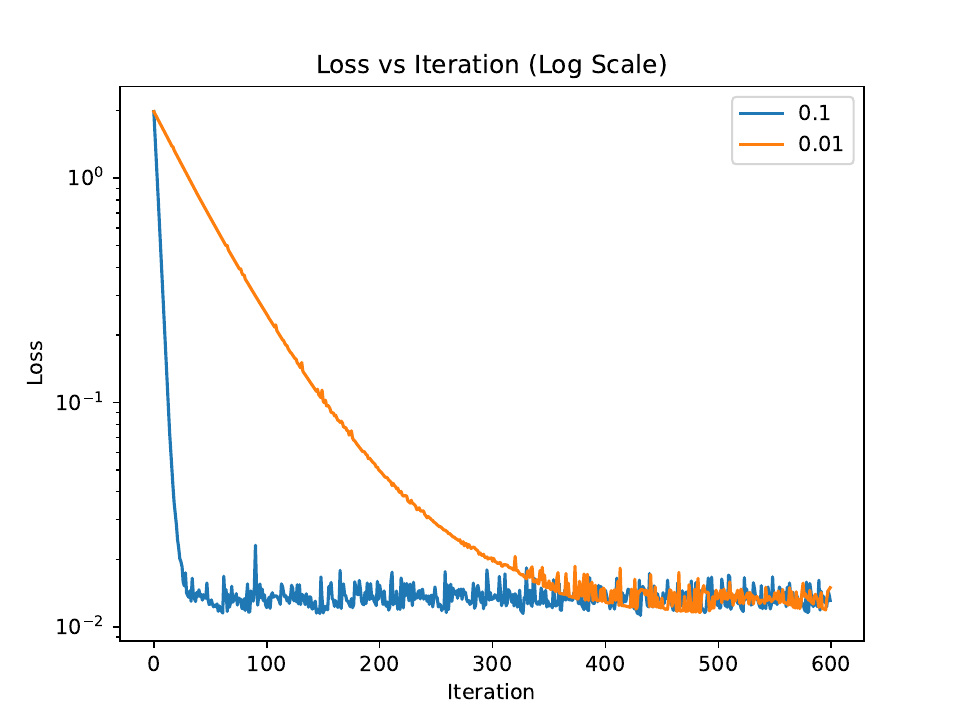}
        \caption{randn}
        \label{fig:randn}
    \end{subfigure}
    \hfill
    \begin{subfigure}{0.49\linewidth}
        \centering
        \includegraphics[width=\linewidth]{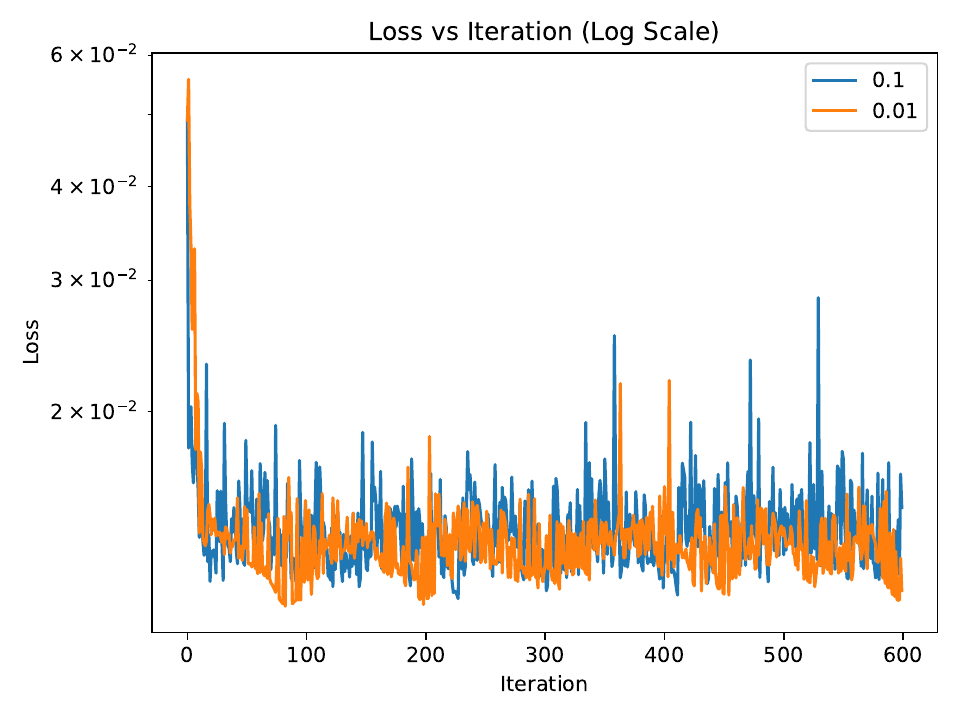}
        \caption{PCA}
        \label{fig:pca}
    \end{subfigure}
    \caption{Comparison of loss curves under different initialization strategies. Subfigure (a) presents the convergence behavior when \(Y\) is initialized with random values (\texttt{randn}), whereas subfigure (b) shows the faster descent typically observed with PCA-based initialization.}

    \label{fig:combined}
\end{figure}

\noindent\textbf{About gradient descent.} To investigate the impact of the learning rate (lr) on the convergence of our method, we conducted a series of tests focused primarily on two values: 0.1 and 0.01. We observed that, for most datasets, using a learning rate of 0.1 led to faster initial convergence while still arriving at final loss values similar to those achieved with lr = 0.01. Figure~\ref{fig:combined} illustrates the loss curves for both lr values (0.1 and 0.01) under two different strategies for generating the set \(Y\). The first strategy initializes \(Y\) with random values drawn from a normal distribution (\texttt{randn}), resulting in no inherent ordering or distance preservation. The second strategy uses PCA, which projects the original dataset to a lower-dimensional space before running gradient descent.

\vspace{1mm}

\begin{figure}[h!]
    \centering
    \begin{subfigure}{0.49\linewidth}
        \centering
        \includegraphics[width=\linewidth]{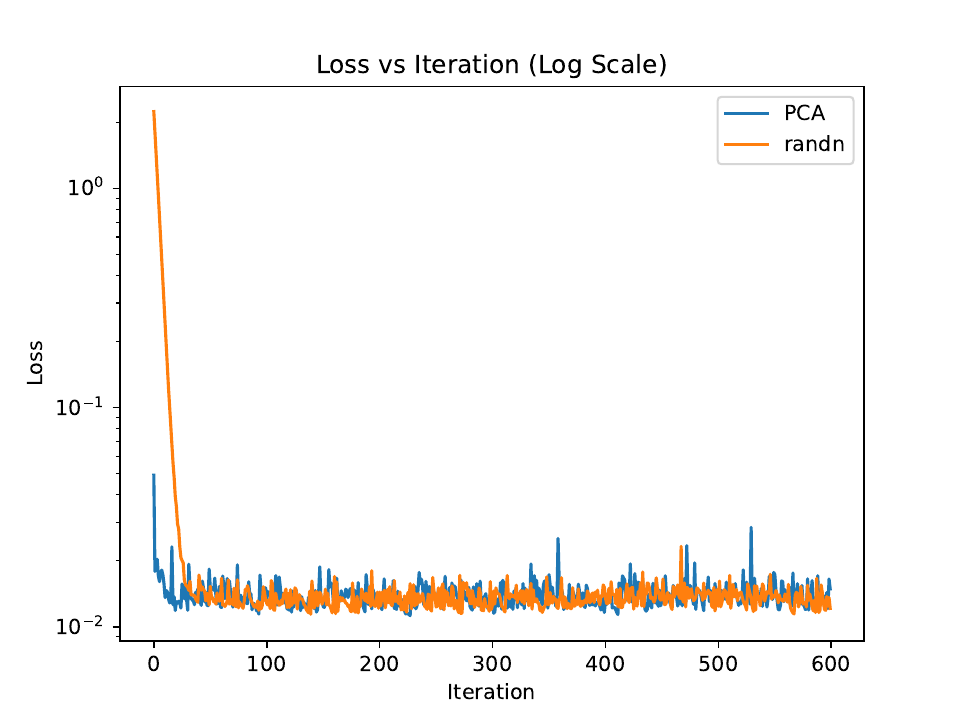}
        \caption{lr = 0.1}
        \label{fig:loss-0.1}
    \end{subfigure}
    \hfill
    \begin{subfigure}{0.49\linewidth}
        \centering
        \includegraphics[width=\linewidth]{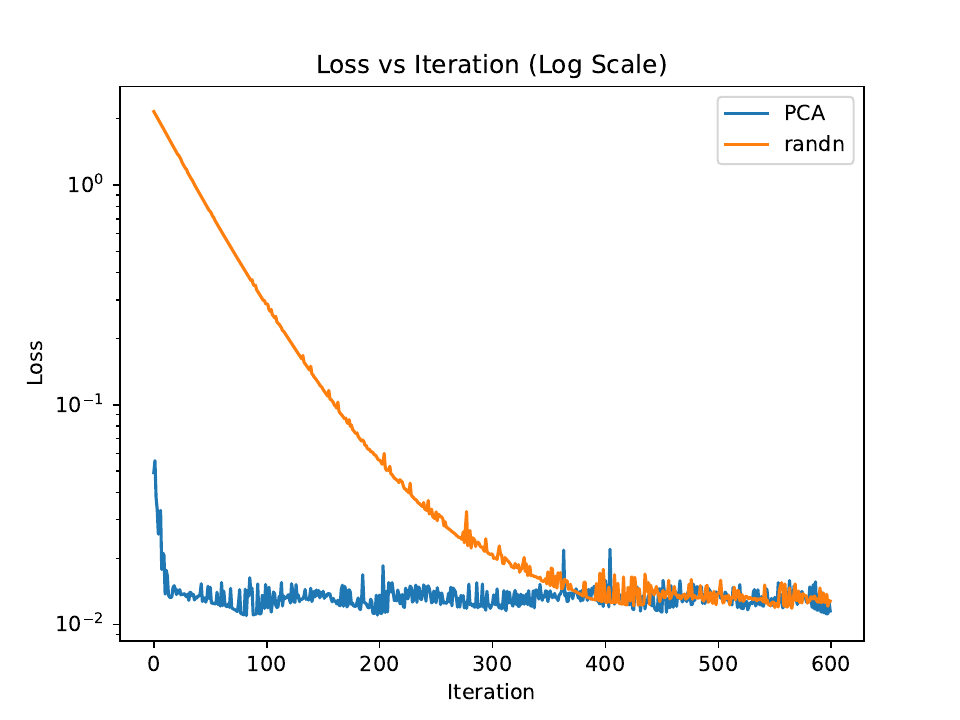}
        \caption{lr = 0.01}
        \label{fig:loss-0.01}
    \end{subfigure}
    \caption{Comparison of the loss curves using two different learning rates (0.1 and 0.01). Subfigure (a) shows the faster initial convergence with lr = 0.1, whereas subfigure (b) illustrates the more gradual descent observed with lr = 0.01. Both rates ultimately converge to similar loss ranges.}

    \label{fig:combined-loss}
\end{figure}

\noindent\textbf{Representation Initialization.}  
The choice of initialization for \(Y\) has a direct impact on both the speed of the algorithm’s convergence. When using \texttt{randn}, each point in the reduced space is placed completely at random, which can require more iterations for the method to “discover” the relevant structure in the data. By contrast, when initializing with PCA, the points already have some level of ordering inherited from the principal components of the original data, even though this projection might not fully preserve distances or capture non-linear relationships.
\vspace{1mm}
In our experiments, we found that PCA initialization often produces lower initial loss values, thus speeding up convergence, as shown in Figure~\ref{fig:combined-loss}. Figures~\ref{fig:loss-0.1} and \ref{fig:loss-0.01} detail the behavior of the loss curves for each learning rate, illustrating how both rates eventually converge to similar ranges after a suitable number of iterations.

\noindent \textbf{Final Note.} Despite these variations, both initialization strategies yield stable embeddings in the end, demonstrating the overall robustness of our approach in adapting to different starting configurations.

\section{Conclusion}\label{sec:conclusion}

In this paper, we introduced a non-linear dimensionality reduction algorithm based on a probabilistic interpretation of data and optimal transport theory. Our experiments with manifold learning datasets and high-dimensional image benchmarks demonstrate the effectiveness of our approach in preserving pairwise distances when embedding points into a lower-dimensional space. In addition, we investigated different optimization strategies such as varying the learning rate and testing alternative initialization methods which highlighted how convergence speed and stability can be significantly improved by fine-tuning these hyperparameters. Future works can explore reducing the computational complexity of our method, for instance, by employing a parametric form for the embedding function or computing the \gls{gw} distance between mini-batches. Furthermore, considering other distance metrics as part of the Gromov-Wasserstein framework could lead to a broader class of algorithms better suited to specific data characteristics.


\bibliographystyle{IEEEbib}
\bibliography{refs}

\end{document}